%% file: main.tex
\documentclass[10pt,twocolumn,letterpaper]{article}

\usepackage{cvpr}              % To produce the CAMERA-READY version
\input{preamble}
\definecolor{cvprblue}{rgb}{0.21,0.49,0.74}
\usepackage[pagebackref,breaklinks,colorlinks,allcolors=cvprblue]{hyperref}
\usepackage[ruled,vlined]{algorithm2e}

\usepackage{booktabs}
\usepackage[table]{xcolor}
\usepackage{colortbl}
\usepackage{array}
\usepackage{multirow}
\usepackage{subcaption}
\usepackage{float}

\usepackage[accsupp]{axessibility}

\usepackage{MnSymbol}
\usepackage{pifont}
\usepackage{amssymb}

\usepackage{comment}

\def\paperID{} % *** Enter the Paper ID here
\def\confName{CVPR}
\def\confYear{2026}

\title{MaskAdapt: Learning Flexible Motion Adaptation via Mask-Invariant Prior \\for Physics-Based Characters}

\author{
Soomin Park \quad
Eunseong Lee \quad
Kwang Bin Lee \quad
Sung-Hee Lee\\
KAIST\\
{\tt\small \href{https://www.soominpark.xyz/MaskAdapt/}{https://www.soominpark.xyz/MaskAdapt}}
}

\begin{document}
\twocolumn[{%
\renewcommand\twocolumn[1][]{#1}%
\maketitle
\vspace{-10pt}
\centering
\includegraphics[width=0.95\linewidth]{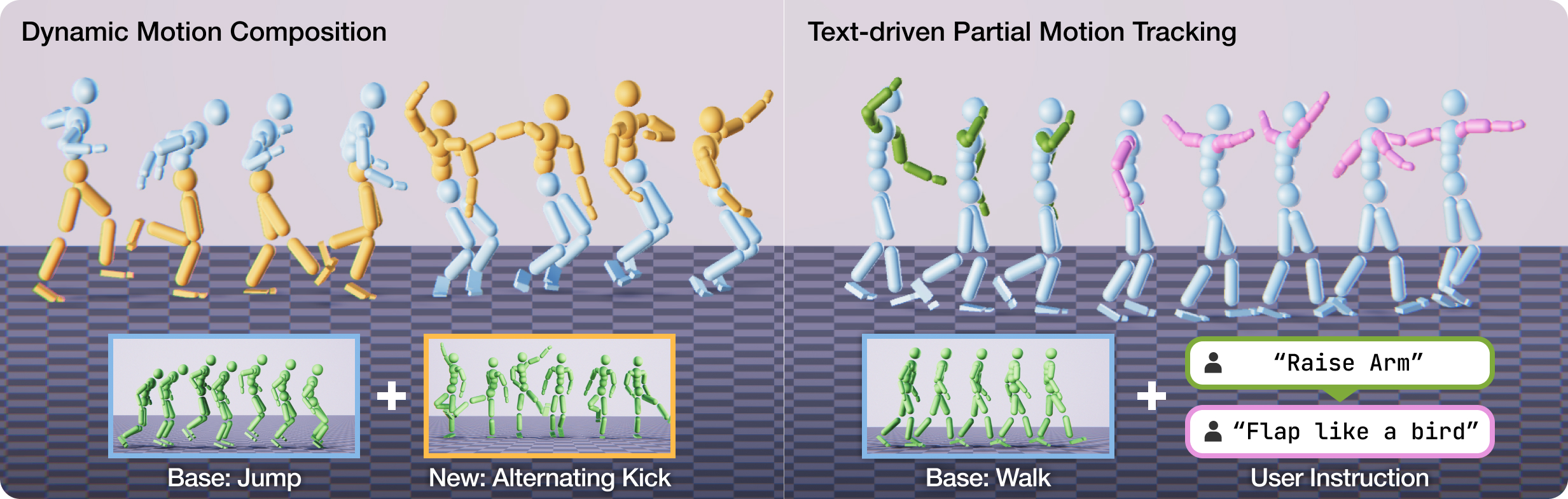}
\captionof{figure}{\textbf{MaskAdapt} enables flexible motion adaptation across two representative tasks: dynamic motion composition, where selected body parts are flexibly modified, and text-driven partial motion tracking, where designated regions follow semantically guided kinematic targets.}
\label{fig:teaser}
}]

% with all behaviors adapted on top of the learned base motion. (a–c) Dynamic motion composition across varying body-part masks, and (d) partial motion tracking of designated body parts (arms) using kinematic targets generated by a text-conditioned model.

\input{sec/0_abstract}   
\input{sec/1_intro}

\input{sec/2_related_work}

\input{sec/3_method}

\input{sec/4_experiment}

\input{sec/6_conclusion}
{
    \small
    \bibliographystyle{ieeenat_fullname}
    \bibliography{main}
}

% WARNING: do not forget to delete the supplementary pages from your submission 
\input{sec/X_suppl}

\end{document}

%% file: sec/0_abstract.tex
\begin{abstract}
We present MaskAdapt, a framework for flexible motion adaptation in physics-based humanoid control.
The framework follows a two-stage residual learning paradigm.
In the first stage, we train a mask-invariant base policy using stochastic body-part masking and a regularization term that enforces consistent action distributions across masking conditions.
This yields a robust motion prior that remains stable under missing observations, anticipating later adaptation in those regions.
In the second stage, a residual policy is trained atop the frozen base controller to modify only the targeted body parts while preserving the original behaviors elsewhere.
We demonstrate the versatility of this design through two applications: (i) motion composition, where varying masks enable multi-part adaptation within a single sequence, and (ii) text-driven partial goal tracking, where designated body parts follow kinematic targets provided by a pre-trained text-conditioned autoregressive motion generator.
Through experiments, MaskAdapt demonstrates strong robustness and adaptability, producing diverse behaviors under masked observations and delivering superior targeted motion adaptation compared to prior work. 
\end{abstract}

%% file: sec/1_intro.tex
%%%%%%%%%%%%%%%%%%%%%%%%%%%%%%%%%%%%%%%%%%%%%%%%%%%%%%%%%%%%%%%%%%%%%%%%%%%%%%
\section{Introduction}
\label{sec:intro}
%%%%%%%%%%%%%%%%%%%%%%%%%%%%%%%%%%%%%%%%%%%%%%%%%%%%%%%%%%%%%%%%%%%%%%%%%%%%%%
% Hook & Recent Work
Creating interactive, physically realistic characters capable of performing a wide range of motions remains a long-standing challenge in both robotics and character animation. 
Recent studies in physics-based humanoid control have sought to extend the skills of pretrained controllers by introducing auxiliary modules while keeping the original weights frozen~\cite{xu2023composite, xu2023adaptnet, xu2024synchronize, won2022physics, pan2025tokenhsi}. These auxiliary modules can take various forms, such as a set of lightweight adaptation layers operating in the latent space~\cite{xu2023adaptnet, xu2024synchronize, pan2025tokenhsi} or a residual architecture~\cite{xu2023composite, won2022physics} whose output is added to that of the base policy. Such designs enable efficient policy reuse and targeted behavioral adaptation without retraining from scratch, while avoiding catastrophic forgetting of the original skills. Among them, residual reinforcement learning (RL) has proven particularly effective for localized motion adaptation, where the residual controller modifies only a subset of body parts while preserving the stability and coordination of the remaining body parts~\cite{xu2023composite}.

% Problem Statement
However, prior residual approaches for motion adaptation face two key limitations. First, they typically train the base policy without accounting for the substantial state-distribution shifts that arise during the subsequent adaptation phase. Consequently, the frozen base controller, never exposed to such variations, often produces unstable actions once the residual policy begins modifying specific body parts. Second, existing composition methods remain limited in flexibility: adaptation is either restricted to fixed body regions or lacks an integrated mechanism for semantic control, such as text-conditioned guidance.

% What's proposed in this paper
To address these limitations, we propose a two-stage framework called MaskAdapt that enables flexible motion adaptation across two representative applications as shown in Fig.~\ref{fig:teaser}.
Our framework comprises two key components: a base policy that learns a robust motion prior, and a residual policy that adapts the motion to target behaviors.

In the first stage, the base policy is trained under stochastic observation masking applied to various body parts and regularized to maintain the consistency in its action distribution with that under full observations. This process encourages the policy to produce consistent actions even when certain inputs are missing, effectively learning to be \textit{mask-invariant}. The resulting motion prior remains robust when masked regions are later modified during residual learning, allowing the policy to rely on visible observations to generate coherent behaviors. 

In the second stage, a residual policy is trained on top of the frozen base policy to modify only the targeted body parts while preserving the rest of the body unchanged. The adapted parts can be either fixed or dynamically varying depending on the scenario. We demonstrate this versatility through two tasks: \emph{Motion Composition} and \emph{Text-Driven Partial Motion Tracking}. In motion composition, users can modify more than one body part within a single sequence, where the adapted motions emerge in those regions according to the provided mask information. In text-driven goal tracking, the adaptable parts are predefined, and free-form textual commands provided at inference control their motion. For this task, we employ a pre-trained autoregressive diffusion model to provide kinematic guidance, from which localized goals around the masked areas are extracted as semantically meaningful targets for the residual controller.
% The user can issue or withdraw commands at any time, allowing the residual controller to seamlessly switch between the base motion and the adapted behaviors.

% Experiments
In the experiments, MaskAdapt demonstrates strong robustness and adaptability. Our base policy exhibits diverse and consistent behaviors under varying masking conditions, and the residual policy, building on this prior, effectively adapts the targeted body parts while preserving the underlying behaviors.
Please also refer to our supplementary video, where our base policy generates diverse motion even under masked observations, comparable to the unmasked baseline~\cite{peng2021amp} and superior adaptation ability over prior work~\cite{xu2023composite}.

%% file: sec/2_related_work.tex
\begin{figure*}
    \centering
    \includegraphics[width=0.95\linewidth]{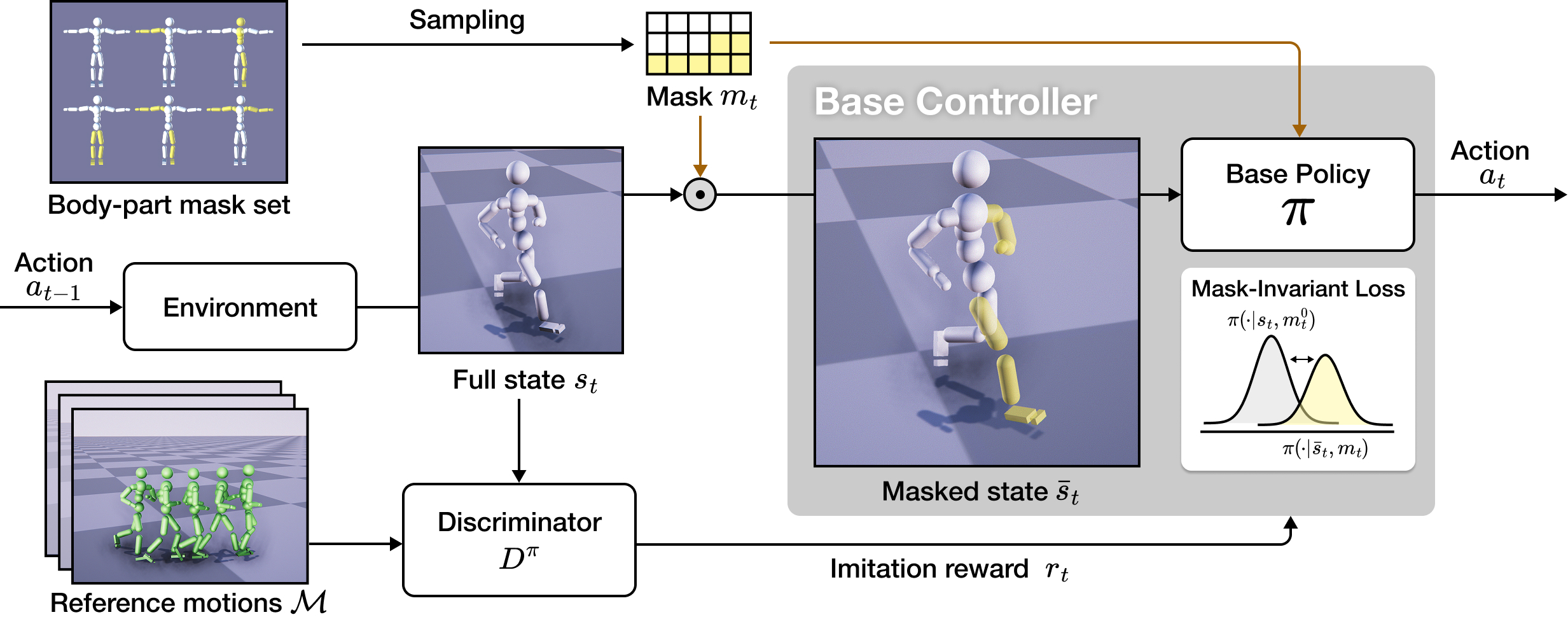}
    \caption{Overview of the base policy training process. At each step, a mask $m_t$ specifies which body parts are hidden from observation. The full state $s_t$ is masked to form $\bar{s}_t$, which is passed to the base policy $\pi$ to generate the action $a_t$. The mask-invariant loss enforces consistency between masked and unmasked action distributions, promoting robustness to partial observations.}
    \label{fig:framework_base}
\end{figure*}

%%%%%%%%%%%%%%%%%%%%%%%%%%%%%%%%%%%%%%%%%%%%%%%%%%%%%%%%%%%%%%%%%%%%%%%%%%%%%%
\section{Related Work}
\label{sec:related}
%%%%%%%%%%%%%%%%%%%%%%%%%%%%%%%%%%%%%%%%%%%%%%%%%%%%%%%%%%%%%%%%%%%%%%%%%%%%%%
Motion adaptation has been a long-standing topic of research, explored across both kinematic and physics-based domains.
In this section, we focus primarily on physics-based approaches, to which our work belongs.

In physics-based control, early work applied generative adversarial imitation learning (GAIL)~\cite{ho2016generative} to train simulated  characters to reproduce complex behaviors from motion capture data.
AMP \cite{peng2021amp} introduced a discriminator-based reward formulation to achieve this goal, followed by ASE \cite{peng2022ase} and CALM \cite{tessler2023calm}, which incorporated encoder- and discriminator-based rewards to map random noise to realistic motions. While these methods produce highly natural movements, they remain constrained by the diversity of motions available in their training datasets.

%%%%%%%%%%%%%%%%%%%%%%%%%%%%%%%%%%%%%%%%%%%%%%%%%%%%%%%%%%%%%%%%
\subsection{Motion Adaptation and Composition}
%%%%%%%%%%%%%%%%%%%%%%%%%%%%%%%%%%%%%%%%%%%%%%%%%%%%%%%%%%%%%%%%

% How can we extend those pretrained controllers to new skills?
%%% Incremental Learning for Adaptation
To extend pretrained controllers beyond their original skill set, recent studies have explored incremental or residual adaptation, where auxiliary network components modify learned behaviors without full retraining~\cite{xu2023composite, xu2023adaptnet, xu2024synchronize, won2022physics, pan2025tokenhsi}. 
AdaptNet~\cite{xu2023adaptnet} introduced auxiliary modules that adjust latent representations for new environments, while PhysicsVAE~\cite{won2022physics} appended a helper branch to refine policy outputs for novel conditions. However, these approaches are largely confined to simple skills and tend to produce only minor stylistic variations. Furthermore, meaningful adaptation typically depends on manually defined task objectives and carefully engineered rewards.

%%% Part-wise Motion Composition
A complementary line of research investigated part-level motion composition to broaden controllable behaviors~\cite{xu2023composite, bae2023pmp, bae2025plt, khoshsiyar2024partwisempc, lee2022learning}.
Composite Motion Learning (CML)~\cite{xu2023composite} employed discriminator ensembles to combine part-wise motions with those generated by a pre-trained policy.
Similarly, PMP~\cite{bae2023pmp} used multiple part-wise discriminators, each providing a distinct learning signal for composing motions of different body regions. PLT~\cite{bae2025plt} utilized a part-wise discrete latent representation updated independently per body part.
Although effective, these architectures require sequential updates across multiple components, resulting in high computational cost and limited scalability. In addition, their adaptation scope is usually restricted to fixed body regions.

Our approach is most closely related to CML, as it also learns a new policy on top of a pre-trained controller. Unlike CML, however, our method allows selective motion composition, enabling adaptation to be applied to different body regions on demand, while preserving the stability of the underlying motion. 
% This design preserves the previously learned skills in the base policy while allowing both reuse of existing behaviors and targeted modification of specific body parts when necessary.

%%%%%%%%%%%%%%%%%%%%%%%%%%%%%%%%%%%%%%%%%%%%%%%%%%%%%%%%%%%%%%%%
\subsection{Masking Mechanisms for Motion Control}
%%%%%%%%%%%%%%%%%%%%%%%%%%%%%%%%%%%%%%%%%%%%%%%%%%%%%%%%%%%%%%%%
% Masking
Several recent works~\cite{tessler2024maskedmimic, pan2025tokenhsi, he2025hover} introduced masking mechanisms to enable controllers to track user-specified targets on demand. Inspired by these efforts, we propose a body-part masking scheme that anticipates which regions will later be adapted in the residual learning stage.
By varying masking conditions during training, the base policy learns stable, mask-invariant motion priors under partial observations, while the residual policy adapts only the masked regions according to localized objectives or semantic goals. 

%%%%%%%%%%%%%%%%%%%%%%%%%%%%%%%%%%%%%%%%%%%%%%%%%%%%%%%%%%%%%%%%
\subsection{Leveraging Text-Driven Kinematic Models}
%%%%%%%%%%%%%%%%%%%%%%%%%%%%%%%%%%%%%%%%%%%%%%%%%%%%%%%%%%%%%%%%
% Integration with text-conditioned kinematic model
Text-guided motion synthesis and editing have recently gained attention as a means for intuitive, high-level control.
Following the success of diffusion-based generative models such as MDM~\cite{tevet2023human}, subsequent works have demonstrated part-wise motion editing through inpainting~\cite{kim2023flame, pinyoanuntapong2024mmm}. More recent methods~\cite{athanasiou2024motionfix, jiang2025dynamic, li2025simmotionedit} further advance fine-grained body-part editing using triplets of source motion, editing instruction, and edited motion.
While these models demonstrate impressive precision in semantic editing, they remain purely kinematic and thus lack physical plausibility.

Hybrid systems such as CLoSD~\cite{tevet2024closd} and DART~\cite{zhao2024dartcontrol} address this issue by generating text-conditioned trajectories that are tracked by physics-based controllers to enforce realism.
However, these approaches limit the controller to a passive tracking role, preventing it from leveraging its own learned dynamics to produce novel, hybrid motions.
In contrast, our framework integrates text-conditioned kinematic guidance directly into the residual learning process, enabling physically consistent yet semantically adaptable control.

%% file: sec/3_method.tex
%%%%%%%%%%%%%%%%%%%%%%%%%%%%%%%%%%%%%%%%%%%%%%%%%%%%%%%%%%%%%%%%%%%%%%%%%%%%%%
\section{Learning Flexible Motion Adaptation}
\label{sec:method}
%%%%%%%%%%%%%%%%%%%%%%%%%%%%%%%%%%%%%%%%%%%%%%%%%%%%%%%%%%%%%%%%%%%%%%%%%%%%%%

%%%%%%%%%%%%%%%%%%%%%%%%%%%%%%%%%%%%%%%%%%%%%%%%%%%%%%%%%%%%%%%%%%%%%%%%%%%%%%
\subsection{Overview}
\label{sec:base_policy}
%%%%%%%%%%%%%%%%%%%%%%%%%%%%%%%%%%%%%%%%%%%%%%%%%%%%%%%%%%%%%%%%%%%%%%%%%%%%%%
We adopt a two-stage framework that first trains a base policy for motion imitation and then learns a residual policy for localized motion adaptation. Throughout both stages, we follow the Adversarial Motion Prior (AMP) framework~\cite{peng2021amp}, where a policy and discriminator are jointly optimized to align generated motions with the reference dataset.

In Sec.~\ref{sec:base_policy}, we describe how the base policy $\pi$ is trained to acquire a robust motion prior. On top of this frozen prior, a residual policy $\psi$ is then learned to modify only selected body parts while leaving the remaining regions unchanged. We demonstrate this capability in two settings: motion composition (Sec.~\ref{sec:motion_composition}), where dynamically chosen body parts are adapted to new behaviors, and text-guided partial goal tracking (Sec.~\ref{sec:partial_tracking}), where designated body regions follow kinematic targets produced by a text-conditioned autoregressive motion generator.

%%%%%%%%%%%%%%%%%%%%%%%%%%%%%%%%%%%%%%%%%%%%%%%%%%%%%%%%%%%%%%%%%%%%%%%%%%%%%%
\subsection{Learning a Mask-Invariant Motion Prior}
\label{sec:base_policy}
%%%%%%%%%%%%%%%%%%%%%%%%%%%%%%%%%%%%%%%%%%%%%%%%%%%%%%%%%%%%%%%%%%%%%%%%%%%%%%
We first train a base policy $\pi$ on the motion dataset $\mathcal{M}$ to establish a robust motion prior for subsequent residual learning. The overall procedure is illustrated in Fig.~\ref{fig:framework_base}. 
The base policy is trained using two key mechanisms: (i) a stochastic observation masking scheme and (ii) a mask-invariant regularization term that encourages consistent actions under different masking conditions.

% Masking scheme
During training, we intentionally mask parts of the state input to expose the policy to imperfect observations. At timestep $t$, the masked state $\bar{s}_t$ is obtained by occluding specific body part-related components of the full state $s_t$ as:
\begin{equation}
\bar{s}_t=(1-m_t) \odot s_t,
\label{eq:mask_state}
\end{equation} 
where $m_t \in \{0,1\}^{|\mathcal{S}|}$ is a binary mask over the state dimension $|\mathcal{S}|$ indicating which elements to zero out. 

Masking operates on functionally coherent joint groups. We partition the full set of joints into $N$ coarse body-part groups $\mathcal{J} = \{\mathcal{G}_1, \dots, \mathcal{G}_N\}$ and a subset of groups $\mathcal{J}_{\text{sub}}$ is sampled with probability $\rho$. 
The mask is then constructed as
\begin{equation}
m_{t,i} =
\begin{cases}
1 & \text{if } f(i)\in\mathcal{J}_{\text{sub}}, \\
0 & \text{otherwise},
\end{cases}
\label{eq:mask_construction}
\end{equation}
where $f:\{0,\dots,|\mathcal{S}|{-}1\}\to\mathcal{J}$ maps each state element to its corresponding body part.
With probability $(1-\rho)$, we apply a null mask, yielding a fully observed state. For clarity, we denote by $m_t^A$ the mask for a subset of body parts $\mathcal{J}_A \subset \mathcal{J}$, 
and by $\bar{s}_t^A$ the resulting masked state.
% This exposes the policy to both complete and partial observations, encouraging robustness to missing inputs. 
The policy receives $(\bar{s}_t, m_t)$ and outputs an action $a_t$, while the discriminator always evaluates \emph{unmasked} transitions $(s, s')$.  
This asymmetric setup forces the policy to generate realistic, expert-like transitions even when observations are corrupted.

% MI loss
While the observation masking scheme enhances robustness through data augmentation, it does not explicitly guarantee consistent behavior across different masking conditions. To address this, we introduce the mask-invariant (MI) loss, a regularization term that explicitly enforces consistency in the policy’s actions under different levels of observation occlusion.
The MI loss is defined as the Kullback–Leibler (KL) divergence between the policy’s action distributions under fully observed and masked states:
\begin{equation}
\mathcal{L}_{\text{MI}} = \mathbb{E}_{m} \left[ \mathrm{D}_\mathrm{KL} \left[\pi(\cdot \mid s, m^0) \,\|\,  \pi(\cdot \mid \bar{s}, m) \right] \right].
\label{eq:mi_loss}
\end{equation}
Here, $\pi(\cdot|s, m^0)$ denotes the action distribution under the full state $s$ (i.e., using the null mask $m^0$) and serves as the consistency target, while $\pi(\cdot \mid \bar{s}, m)$ is the distribution conditioned on the masked observation $\bar{s}$.
Minimizing their divergence encourages the policy to align its behavior under the masked state with that under the full state, thereby making the policy explicitly \emph{mask-invariant}.
The final training objective integrates the MI loss with the PPO objective: $\mathcal{L} = \mathcal{L}_{\text{PPO}} + \lambda_{\text{MI}} \mathcal{L}_{\text{MI}}$, where $\lambda_{\text{MI}}$ controls the strength of the regularization. The training procedure of the base policy is provided in the supplementary material. Once trained, the base policy $\pi$ is frozen for subsequent residual learning.

%%%%%%%%%%%%%%%%%%%%%%%%%%%%%%%%%%%%%%%%%%%%%%%%%%%%%%%%%%%%%%%%%%%%%%%%%%%%%%
\subsection{Dynamic Motion Composition}
\label{sec:motion_composition}
%%%%%%%%%%%%%%%%%%%%%%%%%%%%%%%%%%%%%%%%%%%%%%%%%%%%%%%%%%%%%%%%%%%%%%%%%%%%%%
\begin{figure}
    \centering
    \includegraphics[width=0.92\linewidth]{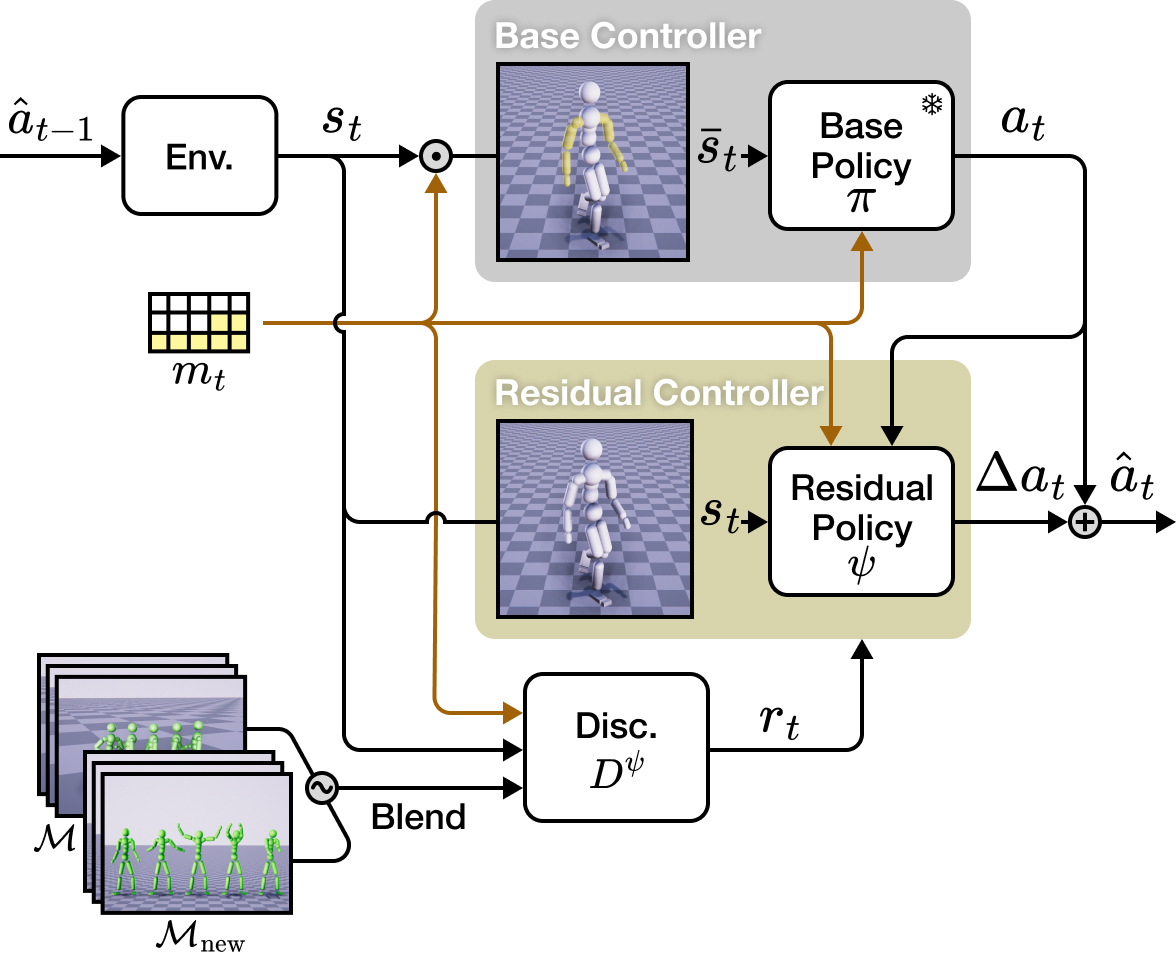}
    \caption{Overview of training the residual policy for motion composition. At each step, a mask $m_t$ specifies which body parts are subject to adaptation, and $\psi$ learns to generate residual actions for those regions while maintaining stability elsewhere.} 
    \label{fig:framework_residual}
\end{figure}

In motion composition, the residual policy $\psi$ leverages the mask information as a conditioning signal, enabling it to selectively adapt the masked body parts to new motions. An overview of this learning process for motion composition is illustrated in Fig.~\ref{fig:framework_residual}. Given the simulated state $s_t$ and the base action $a_t$ produced by $\pi$, the residual policy $\psi$ generates a residual action $\Delta a_t$ to modify only the masked regions while preserving stability in the unmasked parts. The final executed action is obtained by combining the base and residual components:
\begin{equation}
\hat{a}_t = w_{\text{gain}} \cdot \pi(a_t \mid \bar{s}_t, m_t)
          + \psi(\Delta a_t \mid s_t, a_t, m_t),
\label{eq:final_action}
\end{equation}
where $w_{\text{gain}}$ is a gain coefficient applied to $a_t$, implemented either as a learnable parameter or as a fixed value~\cite{xu2023composite}.

Unlike prior approaches~\cite{xu2023composite, bae2023pmp}, which employ an ensemble of discriminators, each dedicated to a particular body part,
we adopt a conditional discriminator $D^{\psi}(s, s' \mid m)$ inspired by CALM~\cite{tessler2023calm}. Whereas CALM conditions on skill embeddings, our discriminator conditions on the mask itself, enabling selective enforcement of motion realism under different masking configurations.
Real samples for the discriminator are constructed kinematically by blending a motion sequence from $\mathcal{M}$ with another sequence from $\mathcal{M}_{\text{new}}$, which contains the new motions to be adapted. We follow the compositing method of~\cite{lee2022learning} for this blending.

Formally, the objective of $D^{\psi}$ is given by:
\begin{equation}
\begin{aligned}
\operatorname*{arg\,max}_{D^{\psi}}
    & \;\; \mathbb{E}_{d^{\mathcal{M}}(s, s')}\!\left[\log D^{\psi}(s, s' \mid m)\right] \\
    & + \mathbb{E}_{d^{\psi}(s, s')}\!\left[\log (1 - D^{\psi}(s, s' \mid m))\right] \\
    & - \lambda_{\text{gp}}\,
        \mathbb{E}_{d^{\mathcal{M}}(s, s')}
        \!\left[
          \bigl\|\nabla_{\theta} D^{\psi}(\theta)\bigr\|_2^{2}
        \right]_{\theta=(s, s', m)} ,
\end{aligned}
\label{eq:cond_disc}
\end{equation}
where $\lambda_{\text{gp}}$ denotes the gradient penalty weight applied to real samples as used in~\cite{peng2021amp, gulrajani2017improved}.

This design enables flexible, body-part–wise motion composition. In Fig.~\ref{fig:results_all}a, the controller first masks the lower body to elicit alternating kicks in the legs, then unmasks the lower body and masks the upper body so the legs revert to jumping while the upper body switches to the adapted motion. In Fig.~\ref{fig:results_all}b, the policy performs fine-grained adaptation by first masking both arms to execute rotating-arms motion while maintaining walking, and then unmasking only the left arm so that the rotation persists in the right arm, yielding smooth, progressive transitions across different body part combinations. Lastly, in Fig.~\ref{fig:results_all}c, the policy remains inactive when no region is masked (exhibiting only the aiming motion), and once the lower body is masked, a sneaking motion emerges in that region, resulting in a natural transition between the base and residual behaviors.

%%%%%%%%%%%%%%%%%%%%%%%%%%%%%%%%%%%%%%%%%%%%%%%%%%%%%%%%%%%%%%%%%%%%%%%%%%%%%%
\subsection{Text-Driven Partial Motion Tracking}
\label{sec:partial_tracking}
\begin{figure}
    \centering
    \includegraphics[width=0.92\linewidth]{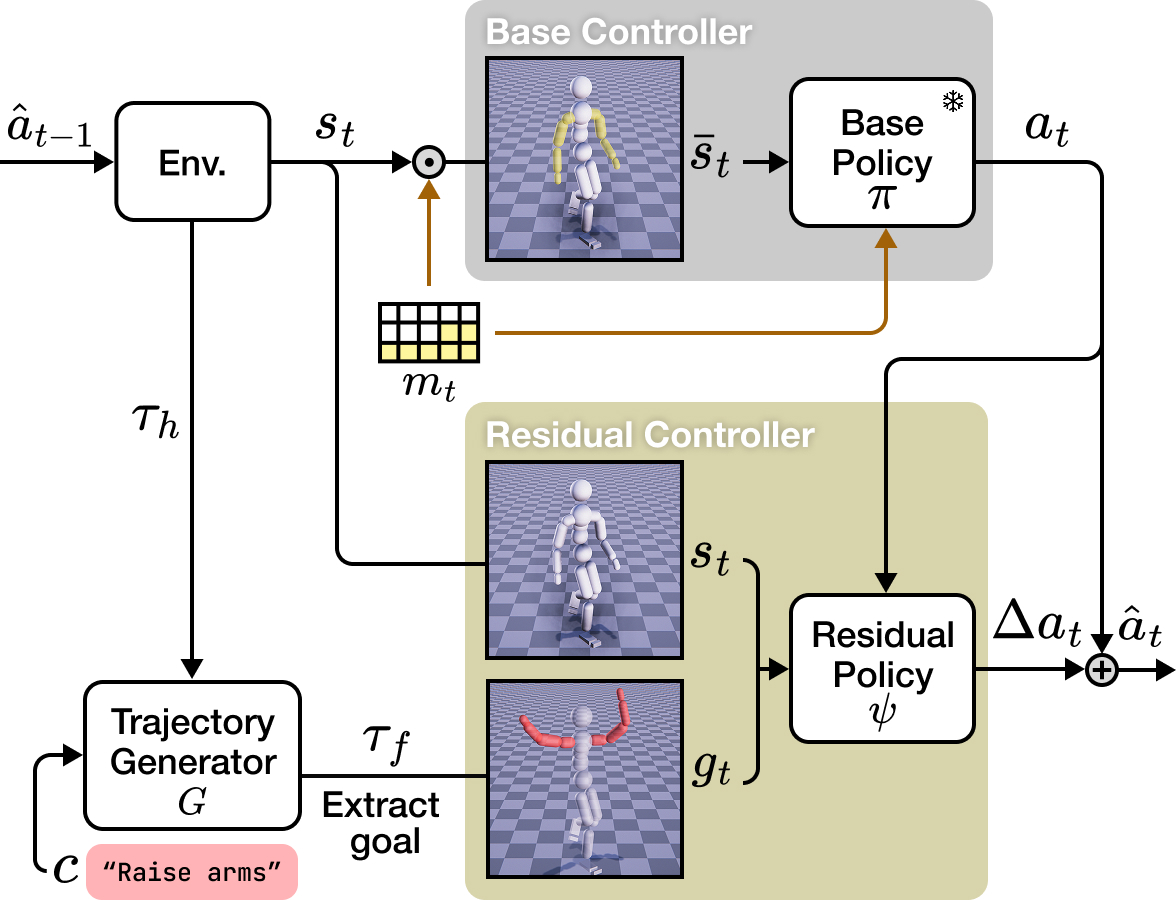}
    \caption{Overview of text-guided partial motion tracking during inference. The trajectory generator $G$ continuously produces kinematic trajectories in response to dynamic text inputs, which serve as tracking targets for the residual policy.}
    \label{fig:framework_residual2}
\end{figure}

%%%%%%%%%%%%%%%%%%%%%%%%%%%%%%%%%%%%%%%%%%%%%%%%%%%%%%%%%%%%%%%%%%%%%%%%%%%%%%
In partial tracking, the residual policy $\psi$ adapts specific body parts to follow motion trajectories defined by text inputs. An overview of this process at inference time is shown in Fig.~\ref{fig:framework_residual2}.
During this stage, a subset of body parts $\mathcal{J}_K$ is designated for modification, and its corresponding mask $m^K$.

To provide kinematic goals, we employ a pre-trained autoregressive diffusion model $G$ from DART~\cite{zhao2024dartcontrol}. Given a motion history
$\tau_h = x_{t-H+1:t}$
and a text command $c$, the generator predicts a short-horizon future sequence $\tau_f$, where localized targets for the designated body parts are extracted as:
\begin{equation}
\tau_f = G(z_T, T, \tau_h, c), \quad
g_t = \text{ExtractGoal}(\tau_f; t, K).
\end{equation}
Here, $z_T \sim \mathcal{N}(0,\mathbf{I})$ is diffusion noise and $T$ is the total number of diffusion steps.
For the selected subset $\mathcal{J}_K$, we construct a text embedding pool from which a text condition $c$ is sampled as input for $G$. 
By repeatedly exposing $\psi$ to kinematic guidance generated from body-part-relevant text conditions, the residual controller learns to adapt to diverse, targeted goals focused on the selected areas. 

\begin{figure*}
    \centering
    \includegraphics[width=0.99\linewidth]{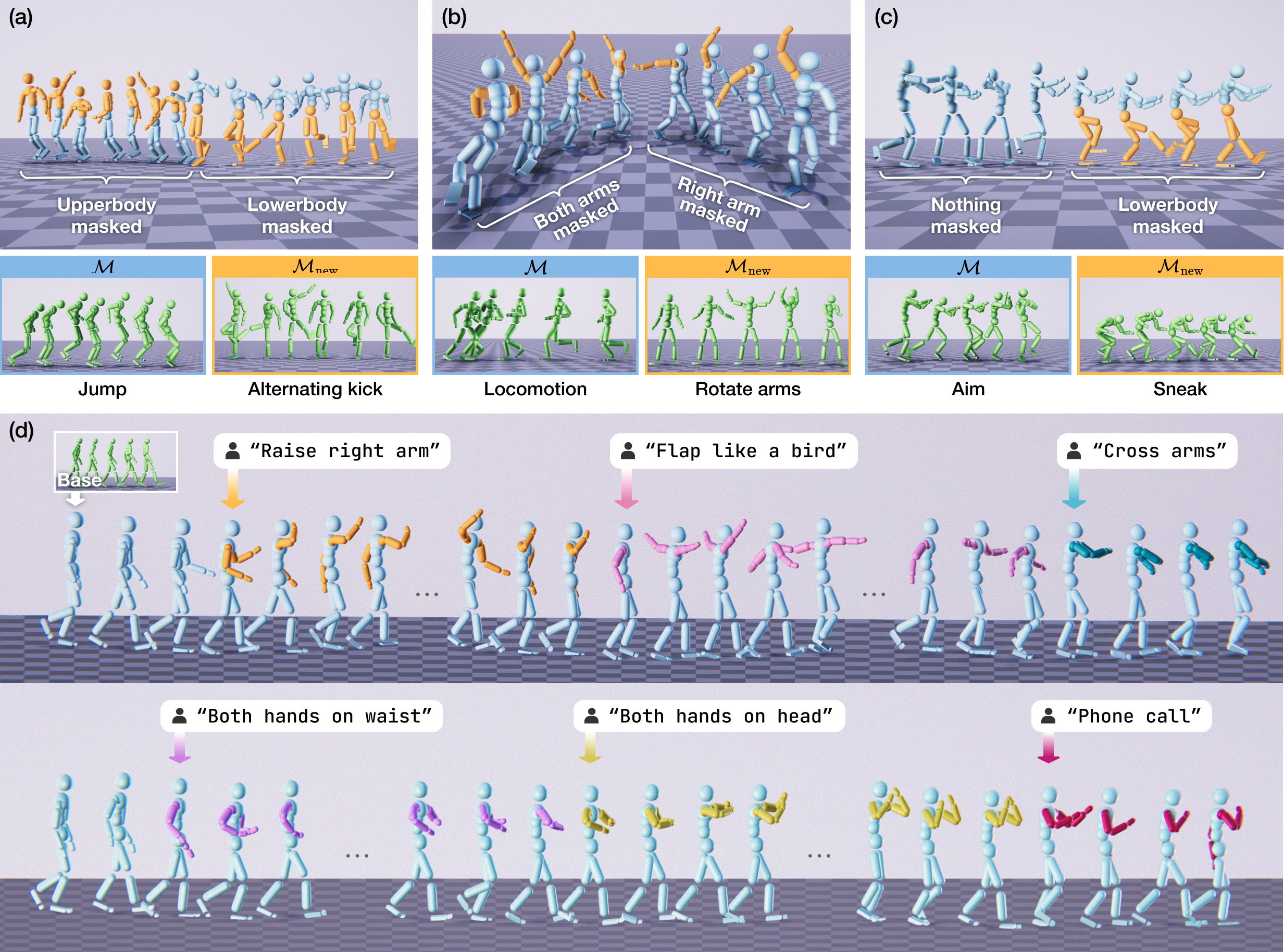}
    \caption{Examples of motion adaptation achieved by the residual policy.
    (a) \textit{Jump} + \textit{Alternating Kick}: Two behaviors are combined via alternating masks on lower and upper body. Kicks first emerge in the legs, then swapping the mask restores jumping while shifting the adapted behavior to the upper body.
    (b) \textit{Locomotion} + \textit{Rotate Arms}: With walking as the base motion, both arms are initially masked to perform rotating-arm motion. Unmasking the left arm yields fine-grained control, where only the right arm continues the adapted behavior, producing smooth, localized transitions.
    (c) \textit{Aim} + \textit{Sneak}: When no region is masked, the controller performs only aiming. Masking the lower body induces sneaking in that region, creating a natural blend between base and adapted behaviors.
    (d) Text-driven partial motion tracking: The character begins with walking and responds to sequential text commands by producing semantically aligned arm motions.}
    \label{fig:results_all}
    
\end{figure*}

% Training vs. Inference
During training, we improve efficiency by periodically pre-generating batches of future trajectories $\tau_f$ with $G$ and storing them in a goal buffer $\mathcal{B}{\tau}$. For each rollout, a trajectory is sampled from $\mathcal{B}{\tau}$, and its goal states $g_t$ are consumed sequentially to guide $\psi$.
At inference time, $G$ runs online: incoming text prompts are immediately converted into future motion predictions, enabling $\psi$ to adapt the designated body parts in real time while the base policy $\pi$ preserves the underlying behavior. This results in interactive text-driven control, as illustrated in Fig.~\ref{fig:results_all}d, where a walking character responds to successive text commands with semantically aligned arm motions. The full training procedure integrating $\psi$ with $G$ is provided in the supplementary material.

\begin{figure*}
    \centering
    \includegraphics[width=0.99\linewidth]{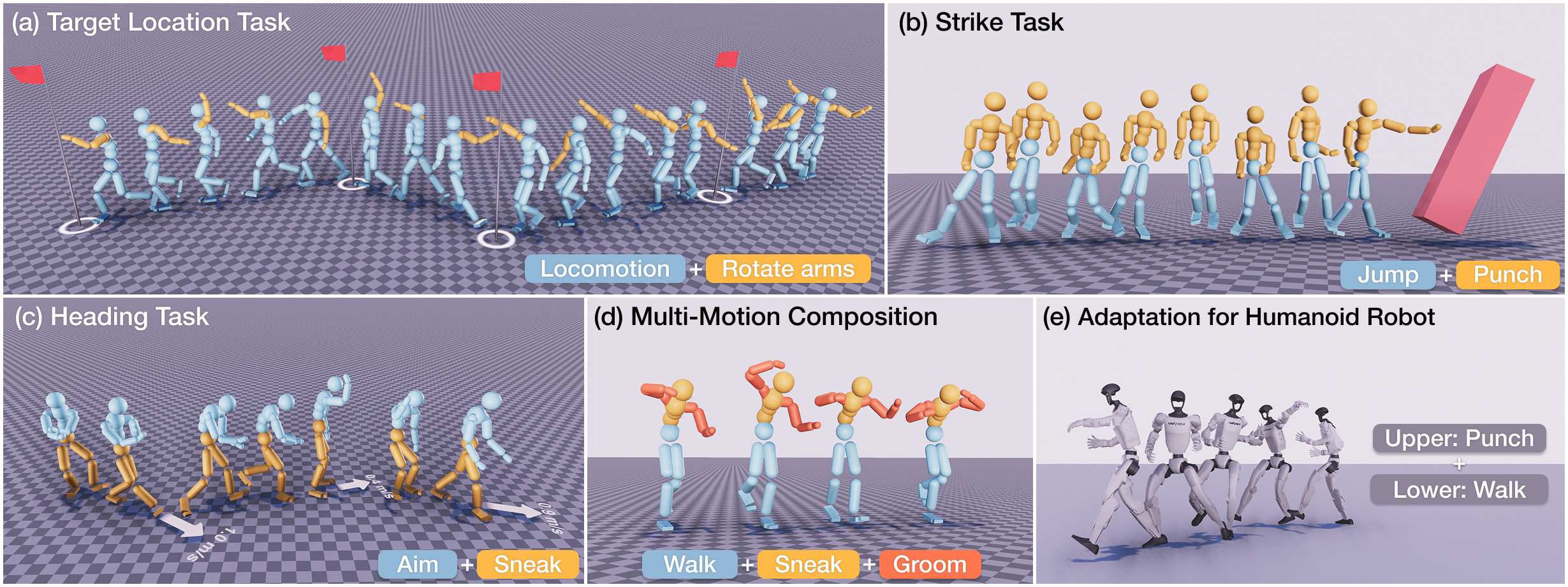}
    \caption{Goal-driven tasks (a--c) and complex scenarios (d--e) enabled by dynamic motion composition.}
    \label{fig:scenarios}
\end{figure*}

%% file: sec/4_experiment.tex
%%%%%%%%%%%%%%%%%%%%%%%%%%%%%%%%%%%%%%%%%%%%%%%%%%%%%%%%%%%%%%%%%%%%%%%%%%%%%%
\section{Experiments}
\label{sec:experiments}
%%%%%%%%%%%%%%%%%%%%%%%%%%%%%%%%%%%%%%%%%%%%%%%%%%%%%%%%%%%%%%%%%%%%%%%%%%%%%%
We evaluate the base policy (Sec.~\ref{sec:eval_base}) as well as the performance of our approach in motion composition (Sec.~\ref{sec:eval_composition}) and partial tracking (Sec.~\ref{sec:eval_tracking}) through both quantitative and qualitative analyses.

\noindent\textbf{Datasets.} 
% Our base policies are trained on motions from LAFAN1~\cite{harvey2020robust}. The motion composition task uses motion clips from AMASS~\cite{AMASS:ICCV:2019}, and the partial tracking task uses trajectories generated by a pre-trained trajectory generator trained on the BABEL~\cite{BABEL:CVPR:2021} dataset.
Our base policies are trained on motions from LAFAN1~\cite{harvey2020robust}. Motion composition uses clips from AMASS~\cite{AMASS:ICCV:2019}, and partial tracking uses trajectories generated by a pre-trained model on BABEL~\cite{BABEL:CVPR:2021}.

\noindent\textbf{Implementation Details.}
We use a simulated humanoid with 22 rigid bodies and 63 degrees of freedom, following the SMPL kinematic structure with the mean body shape from~\cite{luo2023perpetual}. For masking, we partition the 22 joints into five body parts, defining $\mathcal{J}$ = \{\textit{Trunk}, \textit{Left Arm}, \textit{Right Arm}, \textit{Left Leg}, \textit{Right Leg}\}, and allow masking of at most three parts. This granularity follows prior works~\cite{bae2023pmp, athanasiou2023sinc}.
Physics simulation runs in IsaacGym~\cite{makoviychuk2021isaac} at 60Hz, with control at 30Hz via PD controllers tracking target joint positions.
All policy, value, and discriminator networks are implemented as MLPs with hidden dimensions [1024, 1024, 512].  
The discriminator processes 5 consecutive transitions as input, and PPO~\cite{schulman2017proximal} is used for optimization.

%%%%%%%%%%%%%%%%%%%%%%%%%%%%%%%%%%%%%%%%%%%%%%%%%%%%%%%%%%%%%%%%%%%%%%%%%%%%%%
\subsection{Evaluation on Base Policy}
\label{sec:eval_base}
%%%%%%%%%%%%%%%%%%%%%%%%%%%%%%%%%%%%%%%%%%%%%%%%%%%%%%%%%%%%%%%%%%%%%%%%%%%%%%
We aim for a robust base policy that preserves behavioral diversity even under partial observations.
To quantify this, we measure dataset coverage using frame visitation frequency, following the approach of~\cite{zhu2023neural}. 
Frame visitation is computed based on frame-wise similarity between generated motions and the reference motion. For each generated frame, we identify all reference frames with similar joint features within a certain threshold and increment their corresponding bins. Aggregating these counts yields a visitation distribution over the reference motion. 
We then compute the normalized Shannon entropy of this distribution 
$H_{\text{norm}} = -\frac{\sum_{i=1}^n \mathrm{p}_i \log \mathrm{p}_i}{\log n}$,
where $\mathrm{p}_i$ denotes the visitation frequency of the $i$-th frame and $n$ is the total number of reference frames.
This metric captures how closely the method follows the reference and how broadly covers its different motion phases, with values close to $1$ indicating broad coverage and values close to $0$ indicating collapse to a limited set of motion patterns. 

\noindent\textbf{Quantitative Comparison.}
{\sloppy
% We compare our base policy against AMP~\cite{peng2021amp}, trained without any masking or additional loss terms, to evaluate the effect of the mask-invariant (MI) loss. 
We compare our base policy against AMP~\cite{peng2021amp}, which serves as an unmasked baseline trained without observation masking. Our goal is to demonstrate that the our approach can maintain comparable performance even under masking. We train our method both without and with the proposed mask-invariant loss and analyze how each variant performs relative to AMP baseline.
As shown in Tab.~\ref{tab:base_comparison}, AMP achieves the highest entropy value of 0.9124, representing optimal dataset coverage under full observation. When masking is applied without MI loss, entropy drops sharply to 0.7448, indicating strong sensitivity to missing inputs.
In contrast, applying the MI loss ($\lambda_{\text{MI}}{=}1.0$) restores the entropy to 0.9025, nearly matching the unmasked baseline. These results demonstrate that the MI loss effectively prevents collapse under masking and enables the policy to retain dataset-level diversity comparable to the AMP baseline, qualifying it as a robust motion prior.
We further evaluate a weaker regularization setting ($\lambda_{\text{MI}}{=}0.1$), which achieves an entropy of 0.8968. This result indicates that even mild MI regularization substantially improves robustness over the no-MI setting, although it remains slightly below full MI. Additional ablation results are available in the supplementary material (Sec.~\ref{supp:ablation}).
}

\begin{figure}
    \centering    
    \includegraphics[width=0.95\linewidth]{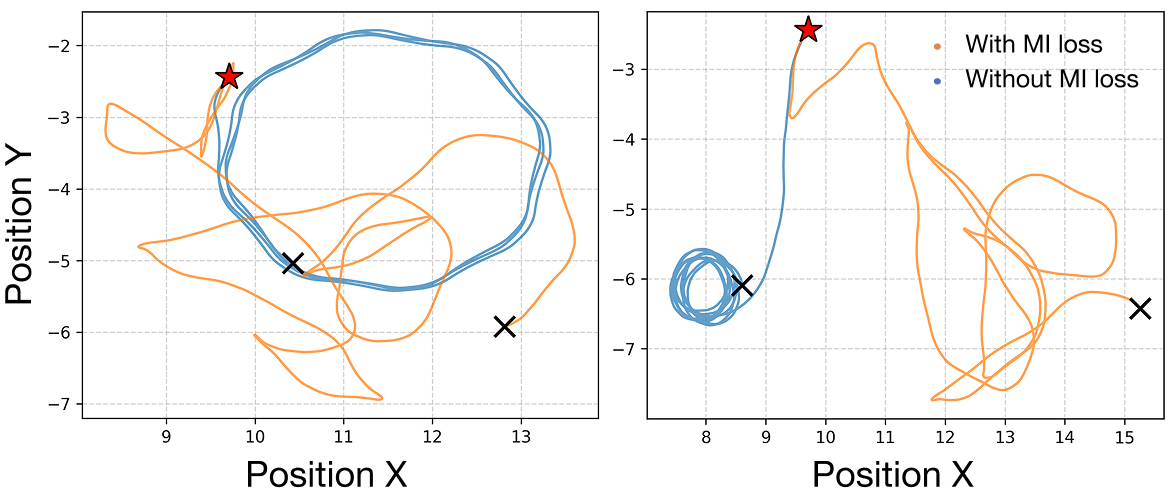}
    \caption{Comparison of root trajectories on the XY plane when both legs (left) and both arms (right) are masked during locomotion, with and without the mask-invariant loss. All sequences start from the same initial state and random seed. The $\medstar$ symbol denotes the starting point, and $\times$ marks the end of each episode.}
    \label{fig:root_trajectories}
\end{figure}

\noindent\textbf{Qualitative Analysis.}
Fig.~\ref{fig:root_trajectories} visualizes root trajectories from 30-second rollout sequences under two extreme masking conditions: masking both legs (left) and both arms (right).
Without the MI loss, motions quickly fall into repetitive loops, producing highly constrained spatial trajectories.
With the MI loss, the trajectories are more dispersed and varied, demonstrating that the base policy can generate diverse and stable locomotion even when critical body parts are masked. Additional results are included in the supplementary video.

{\renewcommand{\arraystretch}{1.0}
\begin{table}[t]
\centering
\small
\setlength{\tabcolsep}{4pt}
\resizebox{\columnwidth}{!}{
\begin{tabular}{p{3.8cm}|p{1.33cm}|p{2cm}}
    \toprule
    Base Policy Model & \centering Masking & \centering $H_{\text{norm}}\uparrow$ \tabularnewline
    \toprule
    \rowcolor{gray!15}
    AMP~\cite{peng2021amp} & \centering $\times$ & \centering 0.9124 \tabularnewline
    \midrule
    $-$ Ours w/o $\mathcal{L}_{\text{MI}}$ & \centering $\checkmark$ & \centering 0.7448 \tabularnewline
    $+$ Ours w/ $\mathcal{L}_{\text{MI}}$ ($\lambda_{\text{MI}} = 0.1$) & \centering $\checkmark$ & \centering 0.8968 \tabularnewline
    $+$ Ours w/ $\mathcal{L}_{\text{MI}}$ ($\lambda_{\text{MI}} = 1.0$) & \centering $\checkmark$ & \centering \textbf{0.9025} \tabularnewline
    \bottomrule
\end{tabular}}
\caption{Normalized entropy ($H_{\text{norm}}$) comparison between the AMP baseline and ablated variants of our base policy trained with and without the MI loss.} 
\label{tab:base_comparison}
\end{table}
}

%%%%%%%%%%%%%%%%%%%%%%%%%%%%%%%%%%%%%%%%%%%%%%%%%%%%%%%%%%%%%%%%%%%%%%%%%%%%%%
\subsection{Evaluation on Motion Composition}
\label{sec:eval_composition}
%%%%%%%%%%%%%%%%%%%%%%%%%%%%%%%%%%%%%%%%%%%%%%%%%%%%%%%%%%%%%%%%%%%%%%%%%%%%%%
In motion composition, body parts controlled by the residual policy are expected to follow the newly introduced motion, while the remaining parts adhere to the original motion learned by the base policy. To evaluate this dual objective, we reuse the normalized entropy metric $H_{\text{norm}}$ introduced earlier, computing it separately for modified and unmodified regions. 
For evaluation, we consider three motion composition cases proposed in Fig.~\ref{fig:results_all}a--c. 
In each case, a single subset of body parts is selected as the adaptation target to simplify the evaluation process. We generate 1,000 frames in 10 parallel environments, yielding 10,000 frames per case.

\noindent\textbf{Quantitative Comparison.}
We compare our method against Composite Motion Learning (CML)~\cite{xu2023composite}, the current state-of-the-art in motion composition.
CML similarly employs a residual controller to modify selected body parts of a base motion, enabling a direct and fair comparison.
Tab.~\ref{tab:comparison_comp} summarizes the results, where each case follows the notation “\textit{A+B (P)},” with \textit{A} as the base motion, \textit{B} as the new motion, and \textit{P} as the adapted body part. Across all cases, our method consistently achieves higher scores for both preserved and adapted regions, indicating stronger retention of the base motion and greater adaptability in the modified parts. Qualitative results are presented in the supplementary material (Sec.~\ref{supp:qualitative_composition}) and the video.

{\renewcommand{\arraystretch}{0.9}
\begin{table}[t]
\small
\centering
\setlength{\tabcolsep}{5pt}
% \small
\begin{tabular}{
    >{\raggedright\arraybackslash}p{1.8cm}|
    >{\centering\arraybackslash\hspace*{-0.15em}}p{0.4cm}|
    @{\hspace{4pt}}
    >{\centering\arraybackslash}p{1.35cm}
    >{\centering\arraybackslash}p{1.35cm}
    >{\centering\arraybackslash}p{1.35cm}
}
\toprule
\multirow{2}{*}{Method} & \multirow{2}{*}{A/B} & \multicolumn{3}{c}{$H_{\text{norm}}\uparrow$} \\
\cline{3-5}

 & \rule{0pt}{2.5ex} & Case (a) & Case (b) & Case (c) \\
\midrule
CML~\cite{xu2023composite} & A & 0.7835 & 0.9193 & 0.8461 \\
                           & B & 0.8788 & 0.7478 & 0.8460 \\
\midrule
Ours                       & A & \textbf{0.8580} & \textbf{0.9268} & \textbf{0.8980} \\
                           & B & \textbf{0.8935} & \textbf{0.8710} & \textbf{0.9214} \\
\bottomrule
\end{tabular}
\caption{Comparison of $H_{\text{norm}}$ across three motion composition cases:
Case definitions: (a) \textit{Jump} + \textit{Alternating Kick} (\textit{Lower Body}),
(b) \textit{Locomotion} + \textit{Rotate Arms} (\textit{Arms}),
(c) \textit{Aim} + \textit{Sneak} (\textit{Lower Body}). Each case is denoted as “\textit{A+B (P)},” with \textit{A} as the base motion, \textit{B} as the new motion, and \textit{P} as the adapted body parts.}
\label{tab:comparison_comp}
\end{table}
}

\noindent\textbf{Goal-driven and Complex Scenarios.}
Our method enables a wide range of goal-driven and complex scenarios through dynamic motion composition.
Fig.~\ref{fig:scenarios}a--c show goal-driven tasks; (a) demonstrates target location task with fine-grained \textit{Rotate Arms} control, indicating that our policy isolates and integrates adapted motions independently of task objectives. (b) shows coordinated execution of an adapted motion (\textit{Punch}) with the base capability (\textit{Jump}) to approach and knock over an object, highlighting effective composition of disparate motor skills. (c) demonstrates concurrent control by following the target heading and speed using \textit{Sneak} while maintaining \textit{Aim}.
Fig.~\ref{fig:scenarios}d shows introducing multiple motions (\textit{Sneak} \& \textit{Groom}) to different body parts (trunk and arms) on top of a base motion (\textit{Walk}).
Finally, we validate our framework on a real humanoid robot (Unitree G1) in Fig.~\ref{fig:scenarios}e, where the policy adapts a walking motion learned by $\pi$ with a punching behavior learned by $\psi$. Please also refer to the supplementary video.

\noindent\textbf{Performance on Goal-Driven Tasks.}
We evaluate performance on goal-driven tasks shown in Fig.~\ref{fig:scenarios}a--c, comparing ours with AdaptNet~\cite{xu2023adaptnet} and CML~\cite{xu2023composite}. Note that AdaptNet does not introduce new motions, but instead modifies base motions to achieve task objectives.
As shown in Tab.~\ref{tab:success_rate}, all methods perform competitively on \textit{Location} and \textit{Heading}, which are less dependent on introducing new motions. In contrast, the \textit{Strike} task, where a new motion (\textit{Punch}) is essential to knock over the box using the wrist, reveals clear performance differences. AdaptNet fails to learn an appropriate wrist-based striking, exploiting alternative body parts (\eg, feet) allowed to make contact. While CML shows improved performance using \textit{Punch}, our method achieves the highest success rate by a significant margin. 
% This result demonstrates that our framework more effectively learn a new motion for goal-directed control, ensuring the faithful execution of the intended motion style.

{\renewcommand{\arraystretch}{0.8}
\begin{table}[t]
\centering
\footnotesize
\resizebox{\columnwidth}{!}{
\begin{tabular}{
    >{\raggedright\arraybackslash}p{2.3cm}|
    @{\hspace{2pt}}
    >{\centering\arraybackslash}p{1.42cm}
    >{\centering\arraybackslash}p{1.4cm}
    >{\centering\arraybackslash}p{1.4cm}
}
\toprule
\multirow{2}{*}{Method} &
\multicolumn{3}{c}{Success Rate$\uparrow$ (\%)} \tabularnewline
\cline{2-4}
& \rule{0pt}{2.3ex}\centering Location & Strike & Heading \tabularnewline
\midrule
AdaptNet~\cite{xu2023adaptnet} & 94.3 & 1.0 & \textbf{82.1 }\tabularnewline
CML~\cite{xu2023composite}      & \textbf{100.0} & 46.5 & 75.2\tabularnewline
Ours     & 99.5 & \textbf{64.2} & 76.9 \tabularnewline
\bottomrule
\end{tabular}}
\caption{Comparison of performance on goal-driven tasks}
\label{tab:success_rate}
\end{table}
}

%%%%%%%%%%%%%%%%%%%%%%%%%%%%%%%%%%%%%%%%%%%%%%%%%%%%%%%%%%%%%%%%%%%%%%%%%%%%%%
\subsection{Evaluation on Partial Goal Tracking}
\label{sec:eval_tracking}
%%%%%%%%%%%%%%%%%%%%%%%%%%%%%%%%%%%%%%%%%%%%%%%%%%%%%%%%%%%%%%%%%%%%%%%%%%%%%%
In partial tracking, body parts controlled by the residual policy follow target trajectories synthesized by $G$, while the remaining parts preserve the base policy’s motion prior (i.e., \textit{walk}). We designate the \textit{Arms} as the adaptable parts. To evaluate these objectives, we measure tracking error between simulated and target poses and DTW-aligned pose error with respect to the reference walking motion.
In our experiments, we use 10 random text prompts, track each sequence for 5 seconds, and repeat three times, yielding 4,500 frames per method.

\begin{figure}
    \centering
    \includegraphics[width=0.95\linewidth]{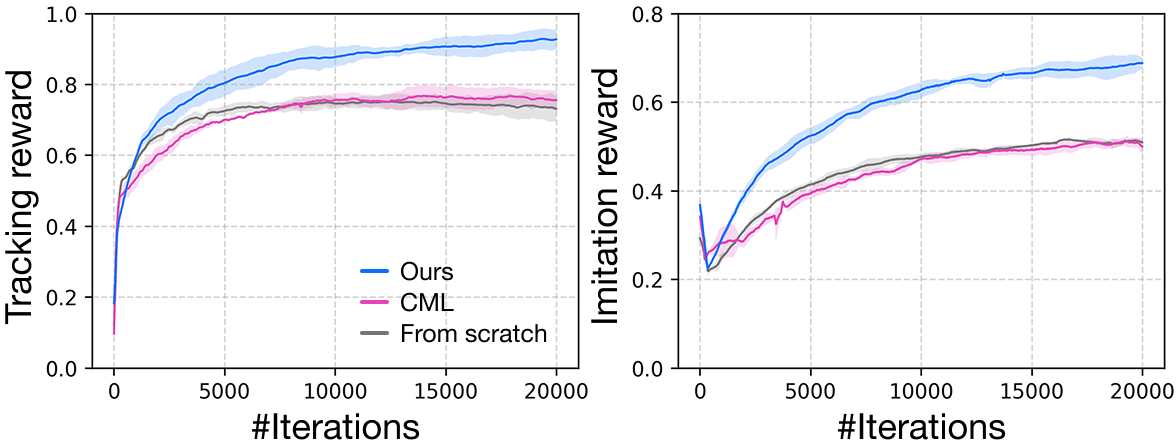}
    \caption{Learning curves of residual policies for tracking (left) and imitation (right) tasks, showing normalized returns averaged over three random seeds.}
    \label{fig:learning_curves}
\end{figure}

\noindent\textbf{Quantitative Comparison.}
We compare our method against CML and a from-scratch model on the partial goal tracking task. Note that the from-scratch model is not built on top of the base policy and therefore lacks an inherent walking prior. For a fair comparison, we adopt the same discriminator setup as CML to preserve base behavior in non-adapted regions.
Tab.~\ref{tab:comparison_track} shows that our method achieves lower errors than both baselines across metrics, indicating more accurate tracking and better preservation of walking motion. Fig.~\ref{fig:learning_curves} reports normalized returns for tracking and imitation, where our method converges faster than models without MI regularization, suggesting more efficient and stable residual learning. Qualitative results are provided in the supplementary material (Sec.~\ref{supp:qualitative_tracking}) and the video.

% \begin{table}[t]
% \centering
% \small
% \begin{tabular}{lcc}
% \toprule
% \textbf{Method} & \textbf{Tracking Error (m)}$\downarrow$ & \textbf{Pose Error (m)}$\downarrow$ \\
% \toprule
% From Scratch & 0.076 $\pm$ 0.005 & 0.057 $\pm$ 0.008\\
% CML~\cite{xu2023composite} & 0.069 $\pm$ 0.001 & 0.079 $\pm$ 0.008\\
% Ours &  \textbf{0.055} $\pm$ 0.007 &  \textbf{0.047} $\pm$ 0.001\\
% \bottomrule
% \end{tabular}
% \caption{Comparison of tracking and imitation errors. Reported as mean $\pm$ standard deviation.}
% \label{tab:comparison_track}
% \end{table}

% {\renewcommand{\arraystretch}{1.0}
\begin{table}[t]
\small
\centering
\resizebox{\columnwidth}{!}{
\begin{tabular}{
    >{\raggedright\arraybackslash}p{2.3cm}|
    c|c
}
\toprule
Method & Tracking Error$\downarrow$ (m) & Pose Error$\downarrow$ (m) \\
\midrule
From Scratch & 0.076$\pm$0.005 & 0.057$\pm$0.008\\
CML~\cite{xu2023composite} & 0.069$\pm$0.001 & 0.079$\pm$0.008\\
Ours &  \textbf{0.055}$\pm$0.007 &  \textbf{0.047}$\pm$0.001\\
\bottomrule
\end{tabular}}
\caption{Comparison of tracking and imitation errors.}
% \caption{Comparison of tracking and pose errors for partial goal tracking.
% Tracking error measures how accurately the policy adapts the designated body parts to the tracking target, while pose error evaluates preservation of the underlying walking motion in the non-adapted body parts. Results are reported as mean $\pm$ standard deviation.}
\label{tab:comparison_track}
\end{table}
% }

\noindent\textbf{Comparison with MaskedMimic.}
% MaskedMimic is similar in spirit to our work in that it composes motions over masked body parts. However, our approach differs in that we adapt new motions onto the masked parts atop a learned behavior, enabling the policy to realize novel combinations of them in both time and space. In this sense, our goal is to expand the motion repertoire of an existing policy, rather than assuming that all possible motion combinations are already encoded within it. % To demonstrate the necessity of our approach, we conduct a quantitative comparison with MaskedMimic (cf. L581–584) on full-body and arm-level tracking, using targets generated from 20 random text prompts. For a fair comparison, MaskedMimic is provided with tracking targets blended with the same walking base behavior for full-body tracking. As shown in Tab.~\ref{tab:comparison_maskedmimic}, our method achieves lower tracking error and higher success rate, while MaskedMimic struggles particularly with arm-only tracking, suggesting that even when trained on large-scale datasets, inpainting motions constrained on specific body parts remains challenging.
We compare with MaskedMimic~\cite{tessler2024maskedmimic} on full-body and arm-level tracking using targets from $G$. For fairness, MaskedMimic is given targets blended with the same base walking motion for full tracking. In Tab.~\ref{tab:comparison_maskedmimic}, our method achieves lower tracking error and higher success rates ($<$0.5m deviation), while MaskedMimic struggles with arm-only tracking, highlighting the challenge of part-constrained motion generation.

{\renewcommand{\arraystretch}{0.9}
\begin{table}
\centering
  {\renewcommand{\arraystretch}{0.82}
   \setlength{\tabcolsep}{2pt}
   \footnotesize
   \resizebox{\columnwidth}{!}{
   \begin{tabular}{
     >{\raggedright\arraybackslash}p{3.2cm}|
    c|c
   }
   \toprule
   Method 
   & Tracking Error$\downarrow$ (m) 
   & Success Rate$\uparrow$ (\%) \\
   \midrule
   MaskedMimic~\cite{tessler2024maskedmimic} (Arms) & 0.0849$\pm$0.003 & 95.5 \\
   MaskedMimic~\cite{tessler2024maskedmimic} (Full) & 0.0699$\pm$0.007 & \textbf{100.0} \\
   Ours (Arms) & \textbf{0.0423}$\pm$0.006 & \textbf{100.0} \\
   \bottomrule
   \end{tabular}}
}
\caption{Comparison of tracking error against MaskedMimic.}
\label{tab:comparison_maskedmimic}
\end{table}
}

%% file: sec/6_conclusion.tex
%%%%%%%%%%%%%%%%%%%%%%%%%%%%%%%%%%%%%%%%%%%%%%%%%%%%%%%%%%%%%%%%%%%%%%%%%%%%%%
\section{Discussion and Conclusion}
\label{sec:conclusion}
%%%%%%%%%%%%%%%%%%%%%%%%%%%%%%%%%%%%%%%%%%%%%%%%%%%%%%%%%%%%%%%%%%%%%%%%%%%%%%
We have presented a framework for flexible motion adaptation, enabled by a mask-invariant training scheme that regularizes the policy to produce consistent actions under masked observations.
The resulting motion prior enables both motion composition and text-driven partial goal tracking. 
Experiments demonstrate that our method achieves robustness comparable to the unmasked baseline while providing superior adaptability over prior work. 

While MaskAdapt demonstrates strong adaptability, several limitations remain that open avenues for future research. Our method may fail when conflicting motions are assigned to closely related body parts (e.g., \textit{Kick} on one leg and \textit{Jump} on the other). Moreover, our text-driven partial tracking relies on a pre-trained kinematic diffusion model, inheriting its semantic constraints and failure modes. Please refer to the supplementary video for details.

\section{Acknowledgments}
\label{sec:acknowledgments}
%%%%%%%%%%%%%%%%%%%%%%%%%%%%%%%%%%%%%%%%%%%%%%%%%%%%%%%%%%%%%%%%%%%%%%%%%%%%%%
We appreciate the anonymous reviewers for their constructive comments that helped improve the final version of this paper. This work was supported by the KOCCA (RS-2025-02307327) and the IITP (RS-2025-25441313), all funded by the Korean government (MCST and MSIT).

%% file: sec/X_suppl.tex
\clearpage
\setcounter{page}{1}
\maketitlesupplementary

\setcounter{section}{0}
\setcounter{figure}{0}
\setcounter{table}{0}

\renewcommand{\thesection}{\Alph{section}}

% To split the supplementary pages from the main paper, you can use \href{https://support.apple.com/en-ca/guide/preview/prvw11793/mac#:~:text=Delete%20a%20page%20from%20a,or%20choose%20Edit%20%3E%20Delete).}{Preview (on macOS)}, \href{https://www.adobe.com/acrobat/how-to/delete-pages-from-pdf.html#:~:text=Choose%20%E2%80%9CTools%E2%80%9D%20%3E%20%E2%80%9COrganize,or%20pages%20from%20the%20file.}{Adobe Acrobat} (on all OSs), as well as \href{https://superuser.com/questions/517986/is-it-possible-to-delete-some-pages-of-a-pdf-document}{command line tools}.

%%%%%%%%%%%%%%%%%%%%%%%%%%%%%%%%%%%%%%%%%%%%%%%%%%%%%%%%%%%%%%%%%%%%%%%%%%%%%%
\section{Implementation Details}
\label{supp:details}
%%%%%%%%%%%%%%%%%%%%%%%%%%%%%%%%%%%%%%%%%%%%%%%%%%%%%%%%%%%%%%%%%%%%%%%%%%%%%%
Base policies were trained for 30K iterations using 4,096 parallel environments. Residual policies were trained for 20K iterations, with 4,096 environments for motion composition and 2,048 for partial tracking. For goal-driven tasks, training was extended to up to 35K iterations.
All policies were optimized with Adam optimizers, using learning rates $1\times10^{-5}$, $1\times10^{-4}$, and $1\times10^{-4}$ for the actor, critic, discriminator, respectively. Other hyperparameters are listed in Tab.~\ref{tab:hyperparams_base} and \ref{tab:hyperparams_residual}.
All experiments were conducted on NVIDIA RTX 4090 GPUs. Training the base policy took about one day, and the residual policy took roughly 16.5 hours for motion composition and 13.5 hours for partial tracking. 

\noindent\textbf{State and Action Representation.}
The state is represented as
$s_t = (\mathbf{h}_t, \mathbf{p}_t,\mathbf{q}_t,\dot{\mathbf{p}}_t,\dot{\mathbf{q}}_t) \in \mathbb{R}^{328}$,
where $\mathbf{h}_t\in \mathbb{R}^1$ represents the root height, $\mathbf{p}_t \in \mathbb{R}^{(J-1) \times 3}$ the joint positions excluding the root, $\mathbf{q}_t \in \mathbb{R}^{J \times 6}$ the joint rotations using 6D representation~\cite{zhou2019continuity}, and $\dot{\mathbf{p}}_t, \dot{\mathbf{q}}_t \in \mathbb{R}^{J \times 3}$ the joint linear and angular velocities, respectively, 
all normalized with respect to the character’s heading transform. Here, $J$ is the total number of joints (in our case 22). The action space matches the number of DoFs, yielding a 63-dimensional vector. 

\noindent\textbf{Reward.} 
For learning the base policy, the imitation reward follows~\cite{peng2021amp}:
\begin{equation}
r_t = -\log(1 - D^{\pi}(s, s')).
\end{equation}
For dynamic motion composition, we use the same formulation but condition the discriminator on the active body-part mask $m$:
\begin{equation}
r_t = -\log\big(1 - D^{\psi}(s, s' \mid m)\big).
\end{equation}
For the partial tracking task, we combine an imitation reward with a tracking reward:
\begin{gather}
r^{\text{imit}}_t = -\log\big(1 - D^{\psi}(\phi_K(s), \phi_K(s'))\big), \\
r^{\text{track}}_t = \exp\left(-\lVert \hat{\mathbf{q}}_t - \mathbf{q}_t \rVert^2 \right),
\end{gather}
where $\phi_K$ extracts the state corresponding to the pre-designated body parts $\mathcal{J}_K$, and $\hat{\mathbf{q}}_t$ denotes the target joint rotations from the generated motion. The final reward is $r_t = w^{\text{imit}} r^{\text{imit}}_t + w^{\text{track}} r^{\text{track}}_t$ with $w^{\text{imit}} = w^{\text{track}} = 0.5$.
The residual policy is optimized through multi-objective learning with PopArt normalization~\cite{xu2023composite, van2016learning} to stabilize learning in the tracking task.

\noindent\textbf{Early Termination.} 
The episode ends when the character falls, determined by any rigid body falling below a height threshold, except for end effectors that are allowed to contact the ground. For partial tracking, we additionally adopt the early termination strategy of~\cite{won2020scalable}, ending the episode when the average reward over a fixed time window becomes sufficiently low.

\begin{table}[t]
\centering
\small
\setlength{\tabcolsep}{6pt}
\begin{tabular}{
    >{\centering\arraybackslash}p{4.0cm}|
    >{\centering\arraybackslash}p{1.7cm}
}
\toprule
Parameters & Value \tabularnewline
\midrule
horizon length                               & 32   \\
number of environments                       & 4096 \\
minibatch size                               & 4096 \\
actor learning rate                          & 1e-5 \\
critic learning rate                         & 1e-4 \\
discriminator learning rate                  & 1e-4 \\
discount factor $\gamma$                     & 0.99 \\
GAE parameter $\tau$                         & 0.95 \\
clip range $\epsilon$                        & 0.2  \\
critic loss weight $\lambda_{V}$             & 5.0  \\
gradient penalty weight $\lambda_\text{GP}$  & 5.0  \\
MI loss weight $\lambda_\text{MI}$           & 1.0  \\
Masking probability $\rho$                   & 0.8  \\
\bottomrule
\end{tabular}
\caption{Hyperparameters for learning base policy.}
\label{tab:hyperparams_base}
\end{table}

\begin{table}[t]
\centering
\small
\setlength{\tabcolsep}{6pt}
\begin{tabular}{
    >{\centering\arraybackslash}p{4.0cm}|
    >{\centering\arraybackslash}p{1.7cm}
}
\toprule
Parameters & Value \tabularnewline
\midrule
horizon length                               & 32   \\
number of environments                       & 4096 / 2048 \\
minibatch size                               & 4096 \\
actor learning rate                          & 1e-5 \\
critic learning rate                         & 1e-4 \\
discriminator learning rate                  & 1e-4 \\
discount factor $\gamma$                     & 0.99 \\
GAE parameter $\tau$                         & 0.95 \\
clip range $\epsilon$                        & 0.2  \\
critic loss weight $\lambda_{V}$             & 5.0  \\
gradient penalty weight $\lambda_\text{GP}$  & 5.0  \\
gain coefficient $w_\text{gain}$             & \textit{learned} \\
\bottomrule
\end{tabular}
\caption{Hyperparameters for learning residual policy.}
\label{tab:hyperparams_residual}
\end{table}

\section{Goal-Driven Tasks}
We describe three goal-driven tasks in detail. The reward function formulation primarily follows prior work~\cite{peng2022ase, xu2023composite}.

\noindent\textbf{Target Location.}
In this task, the objective is to move the character to a target location $\mathbf{p}^{\text{goal}}$.
The task reward is defined as
\begin{equation}
r_t^{g} = \begin{cases}
\exp \left( -3 \frac{\lVert \dot{\mathbf{p}}_t^{\text{root}} - \mathbf{v}_t^* \rVert^2}{\lVert \mathbf{v}_t^* \rVert^2} \right) & \text{if } \lVert \mathbf{p}^{\text{goal}} - \mathbf{p}_t^{\text{root}} \rVert > R, \\
1 & \text{otherwise},
\end{cases}
\end{equation}
where $R$ = 0.5m defines a goal radius around the target location, $\dot{\mathbf{p}}_t^{\text{root}}$ denotes the root velocity, and $\mathbf{v}_t^*$ is the desired velocity toward the goal. The goal observation $g_t \in \mathbb{R}^4$ comprises the relative direction to the target, the target velocity, and the distance to the goal. A trial is considered successful if the character reaches within the $0.5\,\text{m}$ radius of the target location.

\noindent\textbf{Strike Task.}
The objective of character is to approach and strike a box.
The task reward is defined as
\begin{equation}
r_t^{g} = 1 - \mathbf{u}^{\text{world}} \cdot \mathbf{u}^{\text{box}},
\end{equation}
where $\mathbf{u}^{\text{world}} = [0, 0, 1]$ is the world up vector and $\mathbf{u}^{\text{box}}$ is the box's up vector. 
The goal observation $g_t \in \mathbb{R}^{15}$ consists of the state of the target box.  
The success rate is computed based on whether the box tilts beyond $78^\circ$, \ie, $\mathbf{u}^{\text{world}} \cdot \mathbf{u}^{\text{box}} < 0.2$.

\noindent\textbf{Heading Task.}
The heading task requires the character to move in a target direction $\mathbf{d}$ at a specified speed $s^*$, where the target velocity is defined as $\mathbf{v}^* = s^* \mathbf{d}$. 
The reward is formulated as
\begin{equation}
\begin{aligned}
r_t^{g} &=
0.7 \exp \left(-0.25 \left(
\|\mathbf{v}^* - \mathbf{v}_d\|^2
+ 0.1 \|\mathbf{v}_\perp\|^2
\right)\right) \\
&\quad\quad\quad\quad\quad\quad\quad\quad\quad\quad\quad\quad\quad \ + 0.3 \max(\mathbf{d} \cdot \mathbf{f}, 0),
\end{aligned}
\end{equation}
where $\mathbf{v}_d$ is the character's velocity component along the target direction $\mathbf{d}$, $\mathbf{v}_\perp$ is the velocity component orthogonal to $\mathbf{d}$, and $\mathbf{f}$ represents the character's current heading direction. The goal observation $g_t \in \mathbb{R}^3$ comprises the target direction and the target speed. A trial is considered successful if the velocity error along the target direction, $|\mathbf{v}^* - \mathbf{v}_d|$, is less than 0.5.

%%%%%%%%%%%%%%%%%%%%%%%%%%%%%%%%%%%%%%%%%%%%%%%%%%%%%%%%%%%%%%%%%%%%%%%%%%%%%%
\section{Reference Motion Dataset}
\label{supp:dataset}
%%%%%%%%%%%%%%%%%%%%%%%%%%%%%%%%%%%%%%%%%%%%%%%%%%%%%%%%%%%%%%%%%%%%%%%%%%%%%%
Our base policies were trained using motion capture data from LAFAN1~\cite{harvey2020robust}. For the motion composition task, we used motion clips sourced from the AMASS dataset~\cite{AMASS:ICCV:2019}. The partial tracking task relies on trajectories produced by the generator $G$, which are layered on top of the base walking motion, which is also from LAFAN1. A summary of all reference motions used across our experiments is provided in Tab.~\ref{tab:dataset}.

% \begin{table}[t]
% \centering
% \small
% \begin{tabular}{lcrcc}
% \toprule
% \textbf{Motion} & \textbf{Dataset} & \textbf{Duration} & \textbf{Stage}  & \textbf{Task}\\
% \toprule
% Jump & LAFAN1 & 24.4s & Base & C \\
% Locomotion & LAFAN1 & 16m 40s & Base & C\\
% Aim & LAFAN1 & 5m 5s & Base & C\\
% Walk & LAFAN1 & 5.6s & Base & T \\
% Alternating kick & AMASS & 9.9s & Residual & C\\
% Rotate arms & AMASS & 25.13s & Residual & C\\
% Sneak & AMASS & 8.3s & Residual & C\\
% \bottomrule
% \end{tabular}
% \caption{Reference motions used in our experiments. Task type “C’’ denotes motions used for motion composition, and “T’’ indicates motions used for partial tracking.}
% \label{tab:dataset}
% \end{table}

\begin{table}[t]
\centering
\footnotesize
\setlength{\tabcolsep}{4pt}
\begin{tabular}{>{\raggedright\arraybackslash}p{2.0cm}|>{\centering\arraybackslash}p{1.2cm}|>{\raggedleft\arraybackslash}p{1.4cm}|>{\centering\arraybackslash}p{1.2cm}|>{\centering\arraybackslash}p{0.5cm}}
\toprule
\multicolumn{1}{c|}{Motion} & \multicolumn{1}{c|}{Dataset} & \multicolumn{1}{c|}{Duration (s)} & \multicolumn{1}{c|}{Stage} & \multicolumn{1}{c}{Task} \\
\midrule
Jump             & LAFAN1 & 24.4   & Base     & C \\
Locomotion       & LAFAN1 & 1000   & Base     & C \\
Aim              & LAFAN1 & 305    & Base     & C \\
Walk             & LAFAN1 & 5.6    & Base     & T \\
Walk (G1)        & LAFAN1 & 33.3   & Base     & C \\
\midrule
Alternating kick & AMASS  & 9.9    & Residual & C \\
Rotate arms      & AMASS  & 25.1   & Residual & C \\
Sneak            & AMASS  & 8.3    & Residual & C \\
Punch            & AMASS  & 18.3   & Residual & C \\
Punch (G1)       & AMASS  & 14.5   & Residual & C \\
Groom            & AMASS  & 14.5   & Residual & C \\
\bottomrule
\end{tabular}
\caption{Reference motions used in our experiments. Task type "C'' denotes motions used for motion composition, and "T'' indicates motions used for partial tracking.}
\label{tab:dataset}
\end{table}

%%%%%%%%%%%%%%%%%%%%%%%%%%%%%%%%%%%%%%%%%%%%%%%%%%%%%%%%%%%%%%%%%%%%%%%%%%%%%%
\section{Experiments}
\label{supp:experiments}
%%%%%%%%%%%%%%%%%%%%%%%%%%%%%%%%%%%%%%%%%%%%%%%%%%%%%%%%%%%%%%%%%%%%%%%%%%%%%%
%%%%%%%%%%%%%%%%%%%%%%%%%%%%%%%%%%%%%%%%%%%%%%%%%%%%%%%%%%%%%%%%%%%%%%%%%%%%%%
\subsection{Ablation Study on MI Loss Strength}
\label{supp:ablation}
%%%%%%%%%%%%%%%%%%%%%%%%%%%%%%%%%%%%%%%%%%%%%%%%%%%%%%%%%%%%%%%%%%%%%%%%%%%%%%
We conducted an ablation study across a wider range of mask-invariant weights $\lambda_{\text{MI}}$. The full results are presented in Tab.~\ref{tab:mi_ablation}. For each setting, we report the average normalized entropy $H_{\text{norm}}$ computed over all body-part masking configurations. To account for different initialization conditions, we evaluate the policies when motions are generated from random initial states (Random Init.) as well as from a same initial state (Same Init.), and we report $H_{\text{norm}}$ for both cases.
Under the random-initialization setting, a moderate-to-strong MI weight yields the highest entropy, with $\lambda_{\text{MI}} = 5.0$ performing best. Beyond this point, excessively large weights begin to harm performance, reducing entropy even below that of the weakest regularization ($\lambda_{\text{MI}} = 0.1$).
Under the same-initialization setting, the results reflect the policy’s ability to branch into different motions from an identical initial state. In this case, $\lambda_{\text{MI}} = 1.0$ achieves the highest value, outperforming all weaker and stronger regularization strengths. This suggests that a moderate MI weight provides the best balance between enforcing invariance and preserving sufficient freedom for the policy to diverge into a wide range of skills. In our experiments, we adopt $\lambda_{\text{MI}} = 1.0$ as the default setting.

{\renewcommand{\arraystretch}{0.9}
\begin{table*}[t]
\centering
\small
\setlength{\tabcolsep}{6pt}
\begin{tabular}{
    p{3.0cm}|
    >{\centering\arraybackslash}p{1.0cm}|
    >{\centering\arraybackslash}p{2.0cm}|
    >{\centering\arraybackslash}p{2.0cm}
    >{\centering\arraybackslash}p{2.0cm}
}
\toprule
\multirow{2}{*}{Method} &
\multirow{2}{*}{\hspace{-0.15em}Masking} &
\multirow{2}{*}{\parbox{2.0cm}{\centering MI Weights \\ (${\lambda_{\text{MI}}}$)}} &
\multicolumn{2}{c}{Avg.\,$H_{\text{norm}}\uparrow$} \tabularnewline
\cline{4-5}
& & & \rule{0pt}{2.3ex}\centering Random Init. & Same Init. \tabularnewline
\midrule
\rowcolor{gray!15}
AMP~\cite{peng2021amp} & \centering $\times$ & \centering -- & 0.9124 & 0.8890 \tabularnewline
\midrule
$-$ Ours w/o $\mathcal{L}_{\text{MI}}$
    & \centering $\checkmark$ & \centering -- & 0.7448 & 0.7019 \tabularnewline
\midrule
\multirow{7}{*}{$+$ Ours w/ $\mathcal{L}_{\text{MI}}$}
    & \centering $\checkmark$ & \centering 0.1           & 0.8968 & 0.8387 \tabularnewline
    & \centering $\checkmark$ & \centering 0.5           & 0.8987 & 0.8441 \tabularnewline
    & \centering $\checkmark$ & \centering \underline{1.0} (default) & 0.9025 & \textbf{0.8833} \tabularnewline
    & \centering $\checkmark$ & \centering 2.5           & 0.9048 & 0.8387 \tabularnewline
    & \centering $\checkmark$ & \centering 5.0           & \textbf{0.9123} & 0.8552 \tabularnewline
    & \centering $\checkmark$ & \centering 10.0          & 0.9075 & 0.8722 \tabularnewline
    & \centering $\checkmark$ & \centering 50.0          & 0.8803 & 0.8461 \tabularnewline
\bottomrule
\end{tabular}
\caption{Ablation study on the mask-invariant loss weights.}
\label{tab:mi_ablation}
\end{table*}
}

%%%%%%%%%%%%%%%%%%%%%%%%%%%%%%%%%%%%%%%%%%%%%%%%%%%%%%%%%%%%%%%%%%%%%%%%%%%%%%
\subsection{Consistency Across Masking Conditions}
\label{supp:consistency}
%%%%%%%%%%%%%%%%%%%%%%%%%%%%%%%%%%%%%%%%%%%%%%%%%%%%%%%%%%%%%%%%%%%%%%%%%%%%%%
% We further evaluated whether the base policy maintains consistent behavioral diversity under different masking conditions.
% Specifically, we compute $H_{\text{norm}}$ over all masking combinations involving up to two body parts.
% As shown in Tab.~\ref{tab:consistency}, the entropy values vary by only 0.014, indicating that the policy produces similarly diverse motions regardless of which body parts are masked.
% This consistency confirms that the MI-trained base policy serves as a reliable motion prior for subsequent residual learning.

% We also evaluated consistency across all masking conditions without the MI loss. As shown in Tab.~\ref{tab:consistency_2}, $H_{\text{norm}}$ values decrease substantially across all combinations. The policy retains relatively higher values only when a single body part is masked (up to 0.845), but drops further when two parts are masked. The degradation becomes most severe when the masked parts are critical for locomotion. In particular, masking both legs yields the lowest score of 0.602, compared to 0.905 with MI in Tab.~\ref{tab:consistency}. These results highlight the importance of MI regularization in maintaining consistent behaviors under diverse masking configurations.

We further evaluated whether the base policy maintains consistent behavioral diversity under different masking conditions. Specifically, we compute $H_{\text{norm}}$ over all masking combinations involving up to two body parts. As shown in Tab.~\ref{tab:consistency}, the entropy values vary by only 0.014, indicating that the policy produces similarly diverse motions regardless of which body parts are masked. This consistency confirms that the MI-trained base policy serves as a reliable motion prior for subsequent residual learning.

We also evaluated the consistency across all masking conditions when the MI loss is \emph{not} applied. As shown in Tab.~\ref{tab:consistency_2}, the $H_{\text{norm}}$ values decrease substantially across all combinations. The policy retains relatively higher values only when a single body part is masked (up to 0.845), but performance drops further when two body parts are masked. The degradation becomes most severe when the masked parts are critical for locomotion. In particular, masking both legs produces the lowest score of 0.602, showing a large gap compared to the 0.905 achieved when the MI loss is applied in Tab.~\ref{tab:consistency}. These results highlight the importance of MI regularization in maintaining consistent behaviors under diverse masking configurations.

\begin{table}[t]
\centering
\footnotesize
\setlength{\tabcolsep}{4pt}
\renewcommand{\arraystretch}{1.15}
\begin{tabular}{|l|c|c|c|c|c|}
\hline
 & Trunk & Left Arm & Right Arm & Left Leg & Right Leg \\
\hline
Trunk     
& \cellcolor{gray!45}{0.909}
& \cellcolor{gray!30}{0.896}
& \cellcolor{gray!40}{0.906}
& \cellcolor{gray!43}{0.908}
& \cellcolor{gray!42}{0.906} \\
\hline
Left Arm  
& --
& \cellcolor{gray!25}{0.895}
& \cellcolor{gray!33}{0.900}
& \cellcolor{gray!28}{0.898}
& \cellcolor{gray!25}{0.896} \\
\hline
Right Arm 
& --
& --
& \cellcolor{gray!45}{0.907}
& \cellcolor{gray!38}{0.902}
& \cellcolor{gray!37}{0.903} \\
\hline
Left Leg  
& --
& --
& --
& \cellcolor{gray!44}{0.904}
& \cellcolor{gray!44}{0.905} \\
\hline
Right Leg 
& --
& --
& --
& --
& \cellcolor{gray!41}{0.902} \\
\hline
\end{tabular}
\caption{Normalized entropy ($H_{\text{norm}}$) across all masking conditions. Values are arranged in an upper-triangular matrix, with lower entries omitted (“--”) due to symmetry. Diagonal entries correspond to single body-part masking, while off-diagonal entries represent the average $H_{\text{norm}}$ when two body parts are masked simultaneously. Darker shades indicate higher entropy.}
\label{tab:consistency}
\end{table}

\begin{table}[t]
\centering
\footnotesize
\setlength{\tabcolsep}{4pt}
\renewcommand{\arraystretch}{1.15}
\begin{tabular}{|l|c|c|c|c|c|}
\hline
 & Trunk & Left Arm & Right Arm & Left Leg & Right Leg \\
\hline
Trunk     
& \cellcolor{gray!30}{0.798}
& \cellcolor{gray!35}{0.830}
& \cellcolor{gray!28}{0.790}
& \cellcolor{gray!22}{0.749}
& \cellcolor{gray!15}{0.629} \\
\hline
Left Arm  
& --
& \cellcolor{gray!28}{0.814}
& \cellcolor{gray!18}{0.736}
& \cellcolor{gray!25}{0.796}
& \cellcolor{gray!17}{0.702} \\
\hline
Right Arm 
& --
& --
& \cellcolor{gray!35}{0.845}
& \cellcolor{gray!15}{0.681}
& \cellcolor{gray!13}{0.664} \\
\hline
Left Leg  
& --
& --
& --
& \cellcolor{gray!25}{0.754}
& \cellcolor{gray!12}{0.602} \\
\hline
Right Leg 
& --
& --
& --
& --
& \cellcolor{gray!28}{0.783} \\
\hline
\end{tabular}
\caption{Normalized entropy ($H_{\text{norm}}$) across all masking conditions when the MI loss is \emph{not} applied. 
The $H_{\text{norm}}$ values decrease substantially across all combinations, with particularly severe degradation when two body parts are masked simultaneously.}
\label{tab:consistency_2}
\end{table}

%%%%%%%%%%%%%%%%%%%%%%%%%%%%%%%%%%%%%%%%%%%%%%%%%%%%%%%%%%%%%%%%%%%%%%%%%%%%%%
\subsection{Frame Visitation Frequency}
\begin{figure}
    \centering
    \includegraphics[width=1\linewidth]{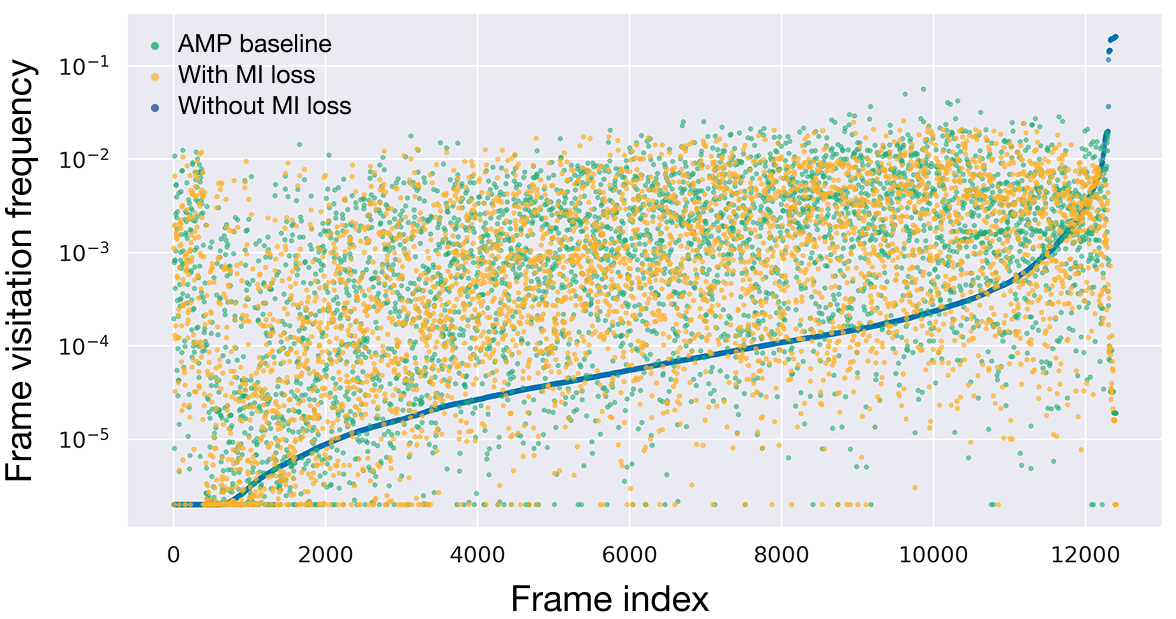}
    \caption{Frame visitation frequencies for locomotion with both legs masked. The policy without MI loss (blue) shows a skewed visitation pattern, whereas applying the MI loss (yellow) produces a markedly more uniform distribution. Our mask-invariant base achieves a level of uniformity comparable to AMP (green), indicating that the MI loss preserves diverse state visitation even under heavy masking.}
    \label{fig:visitation}
\end{figure}

%%%%%%%%%%%%%%%%%%%%%%%%%%%%%%%%%%%%%%%%%%%%%%%%%%%%%%%%%%%%%%%%%%%%%%%%%%%%%%
% Fig.~\ref{fig:visitation} shows the frame visitation distributions for locomotion under an extreme masking condition in which both legs are masked. The y-axis indicates the visitation frequency per frame (log scale), and the x-axis lists frames sorted by their visitation frequency. Without the MI loss (blue), the distribution is highly skewed, revealing that the policy visits only a narrow subset of frames. In contrast, applying the MI loss (yellow) yields a substantially more uniform distribution, indicating that the policy explores a broad range of motion states rather than collapsing onto a small set of highly visited frames.
% For reference, we also show the distribution for the AMP baseline (green), which does not apply any masking strategy. Notably, the base policy with the MI loss achieves a level of uniformity comparable to AMP, demonstrating that it supports robust and diverse state visitation even under heavily masked observations.
Fig.~\ref{fig:visitation} shows frame visitation distributions for locomotion under an extreme masking condition where both legs are masked. The y-axis indicates visitation frequency per frame (log scale), and the x-axis lists frames sorted by visitation frequency. Without the MI loss (blue), the distribution is highly skewed, showing that the policy visits only a narrow subset of frames. In contrast, applying the MI loss (yellow) yields a far more uniform distribution, indicating that the policy explores a broad range of motion states rather than collapsing onto a small set of highly visited frames. For reference, we also show the AMP baseline (green), which does not apply masking. Notably, the base policy with the MI loss achieves a level of uniformity comparable to AMP, demonstrating that it supports robust and diverse state visitation even under heavily masked observations.

%%%%%%%%%%%%%%%%%%%%%%%%%%%%%%%%%%%%%%%%%%%%%%%%%%%%%%%%%%%%%%%%%%%%%%%%%%%%%%
\subsection{Qualitative Comparison for Motion Composition}
\label{supp:qualitative_composition}
\begin{figure}
    \centering
    \includegraphics[width=1\linewidth]{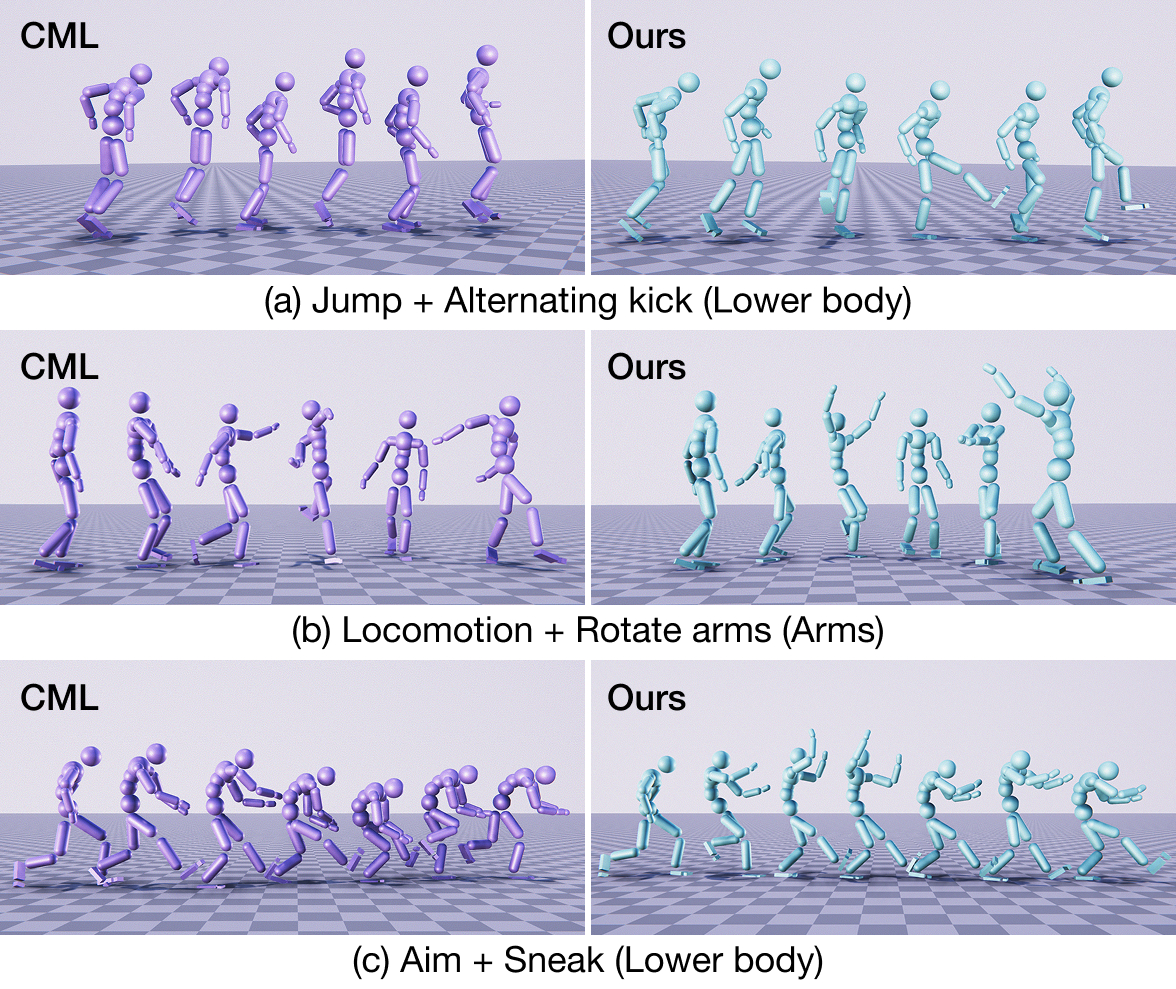}
    \caption{Qualitative comparison of composition methods, each motion generated from the same initial state. In (a), the kick motion in CML is delayed; in (b), CML exhibits severe artifacts and fails to complete full arm rotations; and in (c), CML produces limited motion combinations compared to ours.}
    \label{fig:comparison_comp}
\end{figure}

%%%%%%%%%%%%%%%%%%%%%%%%%%%%%%%%%%%%%%%%%%%%%%%%%%%%%%%%%%%%%%%%%%%%%%%%%%%%%%
Fig.~\ref{fig:comparison_comp} visualizes generated motion sequences for the three composition cases using each method.
Our method produces more natural and coordinated motion combinations, while CML often exhibits delayed activation or incomplete execution of the adapted motion (\eg, limited arm rotation or missing/delayed kick behaviors).
These qualitative results corroborate the entropy-based findings, demonstrating that our framework yields more balanced and expressive motion compositions.
Notably, while CML is constrained to fixed adaptation regions, our method supports more complex scenarios and continues to achieve superior performance under these challenging, high-flexibility settings.

\begin{figure}
    \centering
    \includegraphics[width=1\linewidth]{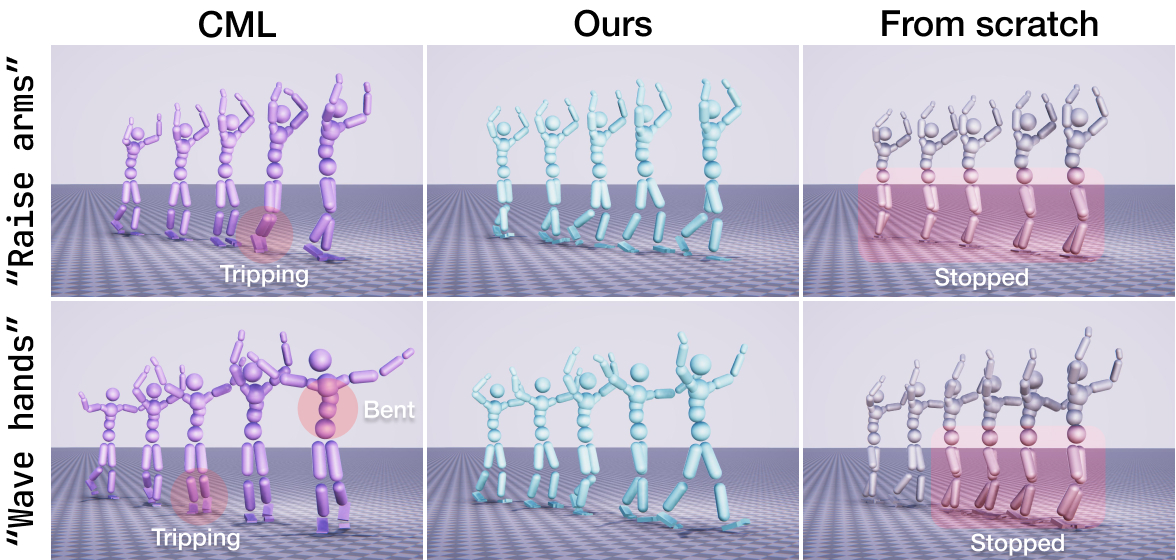}
    \caption{Qualitative comparison of our method, CML, and a from-scratch model for the partial tracking task given the text commands on the left. The red region highlights deviations from the base behavior.}
    \label{fig:comparison_track}
\end{figure}

%%%%%%%%%%%%%%%%%%%%%%%%%%%%%%%%%%%%%%%%%%%%%%%%%%%%%%%%%%%%%%%%%%%%%%%%%%%%%%
\subsection{Qualitative Comparison for Partial Motion Tracking}
\label{supp:qualitative_tracking}
%%%%%%%%%%%%%%%%%%%%%%%%%%%%%%%%%%%%%%%%%%%%%%%%%%%%%%%%%%%%%%%%%%%%%%%%%%%%%%
Fig.~\ref{fig:comparison_track} presents a comparison among our method, CML, and a from-scratch model, all conditioned on the text commands shown on the left. When the character is instructed to execute more pronounced arm motions, such as raising arms or waving hands, both CML and the from-scratch model often struggle to preserve the underlying walk. CML exhibits intermittent disruptions in its stepping pattern, leading to noticeable gait instability. The from-scratch model performs even less reliably and frequently fails to sustain walking. In contrast, our method maintains stable locomotion throughout the sequence. With its improved stability and stronger tracking capability, our controller produces motions that remain coherently aligned with the intended goals provided by the text commands.

%%%%%%%%%%%%%%%%%%%%%%%%%%%%%%%%%%%%%%%%%%%%%%%%%%%%%%%%%%%%%%%%%%%%%%%%%%%%%%
\section{Training Procedure}
%%%%%%%%%%%%%%%%%%%%%%%%%%%%%%%%%%%%%%%%%%%%%%%%%%%%%%%%%%%%%%%%%%%%%%%%%%%%%%
%%%%%%%%%%%%%%%%%%%%%%%%%%%%%%%%%%%%%%%%%%%%%%%%%%%%%%%%%%%%%%%%%%%%%%%%%%%%%%
\subsection{Training Base Policy}
\label{supp:algo_base}
%%%%%%%%%%%%%%%%%%%%%%%%%%%%%%%%%%%%%%%%%%%%%%%%%%%%%%%%%%%%%%%%%%%%%%%%%%%%%%
Algorithm~\ref{algo:base_policy} summarizes the full training procedure for the base policy. For each rollout, we store the transition in the replay buffer $\mathcal{B}$, including the full state $s_t$ and the mask $m_t$ used to construct $\bar{s}_t$. This is necessary since computing the MI loss requires comparing the action distributions under the masked state $\bar{s}_t$ paired with its mask $m^t$ against those under the corresponding full state $s_t$ with the null mask $m^0$.

\input{sec/algo_base_policy}

%%%%%%%%%%%%%%%%%%%%%%%%%%%%%%%%%%%%%%%%%%%%%%%%%%%%%%%%%%%%%%%%%%%%%%%%%%%%%%
\subsection{Training Residual Policy for Partial Tracking}
\label{supp:algo_res}
%%%%%%%%%%%%%%%%%%%%%%%%%%%%%%%%%%%%%%%%%%%%%%%%%%%%%%%%%%%%%%%%%%%%%%%%%%%%%%
Algorithm~\ref{algo:residual_policy} outlines the training procedure for the residual policy in the partial motion tracking task. 
To improve efficiency, we sampled batches of future trajectories $\tau_f$ every $N_{\text{sample}}$ iterations using $G$ and store them in a goal buffer $\mathcal{B}_{\tau}$. 
For each rollout, a trajectory was sampled from $\mathcal{B}_{\tau}$, and its future goal states $g_t$ were consumed sequentially to guide the residual policy $\psi$.

\input{sec/algo_residual_policy}

%%%%%%%%%%%%%%%%%%%%%%%%%%%%%%%%%%%%%%%%%%%%%%%%%%%%%%%%%%%%%%%%%%%%%%%%%%%%%%
\section{Details on Trajectory Generator}
%%%%%%%%%%%%%%%%%%%%%%%%%%%%%%%%%%%%%%%%%%%%%%%%%%%%%%%%%%%%%%%%%%%%%%%%%%%%%%
The generator $G$ is a diffusion-based autoregressive motion generation model~\cite{zhao2024dartcontrol} that takes a text prompt $c$ and a history motion $\tau_h = x_{t-H+1:t}$ as conditioning inputs, and predicts a future motion segment $\tau_f = x_{t+1:t+F}$.

\noindent\textbf{Motion Representation.} Each pose in the sequence is represented as
$x_t = (\mathbf{t}_t,\, \mathbf{o}_t,\, \mathbf{p}_t,\, \mathbf{\Theta}_t,\,
\dot{\mathbf{t}}_t,\, \dot{\mathbf{o}}_t,\, \dot{\mathbf{p}}_t),$
where $\mathbf{t}_t \in \mathbb{R}^3$ is the root translation, 
$\mathbf{o}_t \in \mathbb{R}^6$ the root orientation in 6D representation, 
$\mathbf{\Theta}_t \in \mathbb{R}^{(J-1)\times 6}$ the local joint rotations, 
and $\mathbf{p}_t \in \mathbb{R}^{J\times 3}$ the joint positions. 
The dotted terms represent the first-order temporal derivatives (velocities) of the corresponding quantities. The motion is canonicalized with respect to the first frame of the history motion.

\noindent\textbf{Formulation.}
The generator $G$ is implemented as a latent diffusion model~\cite{chen2023executing}, where generation operates within the compact latent space constructed by the VAE. In the forward diffusion process, Gaussian noise is progressively added to the latent variable $z_0$ according to
\begin{equation}
q(z_k \mid z_{k-1}) =
\mathcal{N}\!\left(\sqrt{\alpha_k}\, z_{k-1},\, (1-\alpha_k) \mathbf{I}\right),
\end{equation}
where $\alpha_k \in (0, 1)$ is a hyperparameter for noise scheduling at diffusion step $k$.
The diffusion model is trained to approximate the reverse process
\begin{equation}
p_\theta(z_{k-1} \mid z_k,\, k,\, \tau_h,\, c)
= \mathcal{N}(\mu_\theta,\, \tilde{\sigma}_k\mathbf{I}),
\end{equation}
where $\mu_\theta$ is modeled by a neural network parameterized by $\theta$, and $\tilde{\sigma}_k$ is the variance specified by the noise schedule.
During generation, the model iteratively denoises $z_k$ from $k = T$ down to $0$, and the final latent vector $\hat{z}_0$ is decoded by the VAE decoder to obtain the clean future motion $\tau_f$. For convenience, we refer to the entire sampling pipeline as $\tau_f = G(z_T, T, \tau_h, c)$.

\noindent\textbf{Implementation details.} We used a history length of $H=5$ and a future length of $F=15$. The total number of diffusion steps was $T=10$. Both the VAE and the diffusion model were implemented using Transformer architectures. We performed sampling using the DDPM procedure and applied classifier-free guidance (CFG) with a guidance weight of 5.0 for text-conditioned generation. The text input was encoded using the CLIP~\cite{radford2021learning} text encoder. Following DART~\cite{zhao2024dartcontrol}, we trained the diffusion model using a simple objective, where the model directly predicts the clean latent code rather than the added noise.

%%%%%%%%%%%%%%%%%%%%%%%%%%%%%%%%%%%%%%%%%%%%%%%%%%%%%%%%%%%%%%%%%%%%%%%%%%%%%%
\section{Details on Text Embedding Pool}
\label{sec:text_set}
%%%%%%%%%%%%%%%%%%%%%%%%%%%%%%%%%%%%%%%%%%%%%%%%%%%%%%%%%%%%%%%%%%%%%%%%%%%%%%
In partial tracking task, we defined a subset of body parts $\mathcal{J}_K$ for adaptation and sample a text input $c$ from its associated embedding pool $\mathcal{C}$ as input to $G$. 
To construct $\mathcal{C}$, we used OpenAI’s GPT-4.1 API to automatically determine the body-part involvement of each action label in the BABEL dataset~\cite{BABEL:CVPR:2021}, following a similar strategy to \cite{athanasiou2023sinc}. 
GPT is guided through three sequential steps:
(1) Action clarity check: Determine whether the label contains enough information to infer body-part involvement. Vague labels are skipped with a brief justification.
(2) Body-part analysis: For analyzable labels, identify the essential body parts based on an isolated recognizability criterion, meaning a part is chosen only if the action remains identifiable when observed in isolation.
(3) Structured output: Return a JSON entry containing the selected parts and a brief rationale, or a skip flag with a skip reason.
% For example, given an action label (\eg, “clap”), GPT is instructed to identify the most relevant body part (\eg, \emph{Arms}). Each query includes both the raw action label and its associated action categories (\eg, “hand movement”) to reduce potential ambiguity. 
After the body-part assignments are obtained, we extracted only the descriptions associated with the designated parts to construct $\mathcal{C}$. In our experiments, we constructed $\mathcal{C}$ for the \emph{Arms} region and inputted arms-related texts as conditioning inputs to $G$ to produce the kinematic guidance.

%%%%%%%%%%%%%%%%%%%%%%%%%%%%%%%%%%%%%%%%%%%%%%%%%%%%%%%%%%%%%%%%%%%%%%%%%%%%%%
\section{Failure Modes}
\label{sec:failure_modes}
%%%%%%%%%%%%%%%%%%%%%%%%%%%%%%%%%%%%%%%%%%%%%%%%%%%%%%%%%%%%%%%%%%%%%%%%%%%%%%
Our method may struggle when conflicting motions are assigned to closely related body parts. For example, combining a \textit{Kick} on one leg with a \textit{Jump} on the other can produce unnatural motion. This is because motions like jumping inherently require coordinated support from multiple related parts, making them difficult to realize when one part is constrained by a different objective.

%% file: sec/algo_base_policy.tex
\begin{algorithm}[t]
\small
\caption{Training Base Policy}
\label{algo:base_policy}
\SetAlgoLined
\DontPrintSemicolon

\BlankLine
Initialize base policy $\pi$ and value function $V^{\pi}$\;
Initialize discriminator $D^{\pi}$\;
Initialize replay buffer $\mathcal{B}$\;

\BlankLine

\While{training not converged}{
  \tcc{Collect rollouts}
  \For{$t = 1$, ...,  $L$}{
    Sample mask $m_t$\;
    Compute masked state $\bar{s}_t$ from $(s_t, m_t)$ using Eq.~\eqref{eq:mask_state} \;
    Sample action $a_t \sim \pi(a_t \mid \bar{s}_t, m_t)$\;
    Simulate next state $s_{t+1}$ using $s_t$ and $a_t$\;
    Store transition $(s_t, m_t, a_t, s_{t+1})$ in buffer $\mathcal{B}$\;
    $s_t \gets s_{t+1}$\;
  }

  \BlankLine
  \tcc{Update discriminator}
  Update $D^{\pi}$ on $(s, s')$ from $\mathcal{M}$ and $\mathcal{B}$\;

  \BlankLine
  \tcc{Update base policy}
  \For{$u = 1,..., n$}{
    Sample $(s_t, m_t, a_t, s_{t+1})$ from $\mathcal{B}$\;
    % Sample another mask $m_t' \neq m_t$ and compute $\bar{s}_t'$\;
    Compute $\mathcal{L}_{\text{MI}}$ from $(s_t, m^0, m_t)$ using Eq.~\eqref{eq:mi_loss} \;
    $\mathcal{L} \gets \mathcal{L}_{\text{PPO}} + \lambda_{\text{MI}}$ \;
    Update $\pi$ and $V^{\pi}$ with $\mathcal{L}$\;
  }
}
\end{algorithm}

%% file: sec/algo_residual_policy.tex
\begin{algorithm}[t]
\small
\caption{Training Residual Policy for Partial Tracking}
\SetAlgoLined
\DontPrintSemicolon

$\pi$ : pre-trained base policy \;
$G$ : pre-trained trajectory generator \;
$\mathcal{J}_K$ : pre-defined subset of body parts \;
$\mathcal{C}$ : text embedding pool for $\mathcal{J}_K$ \;
% $\mathcal{B}_{\tau}$ : trajectory buffer \;
% $\mathcal{B}$ : replay buffer \;

\BlankLine
Initialize residual policy $\psi$ and value function $V^\psi$ \;
Initialize discriminator $D^{\psi}$\;

\BlankLine

\While{training not converged}{
    \tcc{Sample trajectories offline}
    \If{$i \bmod N_{\text{sample}} = 0$}{
        Sample random noise $z_T \sim \mathcal{N}(\mathbf{0}, \mathbf{I})$ \;
        Sample text input $c \sim \mathcal{C}$ \;
        Sample $\tau_f \sim G(z_T, T, \tau_h, c)$\;
        Store $\tau_f$ in $\mathcal{B}_\tau$ \;
    }
    
    \BlankLine
    \tcc{Collect rollouts}
    Sample $\tau_f \sim \mathcal{B}_\tau$ \;
    \For{timesteps $t = 1, \dots, F$}{
        Compute $\bar{s}_t^K$ from $(s_t, m_t^K)$ using Eq.~\eqref{eq:mask_state} \;
        Compute $g_t \gets \text{ExtractGoal}(\tau_f; t, K)$ \;
        Sample $a_t \sim \pi(a_t|\bar{s}_t^K, m^K)$ \;
        Sample $\Delta a_t \sim \psi(\Delta a_t|s_t, g_t, a_t)$ \;
        Compute $\hat{a}_t = w_{\text{gain}} \cdot a_t + \Delta a_t$\;
        Simulate next state $s_{t+1}$ using $s_t$ and $\hat{a}_t$ \;
        Store transition $(s_t, g_t, m_t^K, \hat{a}_t, s_{t+1})$ in $\mathcal{B}$\;
        $s_t \gets s_{t+1}$ \;
    }
    \BlankLine
    \tcc{Update discriminator}
    Update $D^{\pi}$ on $(s, s')$ from $\mathcal{M}$ and $\mathcal{B}$\;
    
    \BlankLine
    \tcc{Update residual policy}
    Update $\psi$ and $V^\psi$ using data from $\mathcal{B}$\;
}
\label{algo:residual_policy}
\end{algorithm}

%% file: main.bib
@String(CVPR= {IEEE Conf. Comput. Vis. Pattern Recog.})

@String(TOG= {ACM Trans. Graph.})

@String(AAAI = {AAAI})

@String(CVPR  = {CVPR})

@String(TOG   = {ACM TOG})

@inproceedings{athanasiou2024motionfix,
  title={Motionfix: Text-driven 3d human motion editing},
  author={Athanasiou, Nikos and Cseke, Alp{\'a}r and Diomataris, Markos and Black, Michael J and Varol, G{\"u}l},
  booktitle={SIGGRAPH Asia 2024 Conference Papers},
  pages={1--11},
  year={2024}
}

@inproceedings{
tevet2023human,
title={Human Motion Diffusion Model},
author={Guy Tevet and Sigal Raab and Brian Gordon and Yoni Shafir and Daniel Cohen-or and Amit Haim Bermano},
booktitle={The Eleventh International Conference on Learning Representations },
year={2023},
url={https://openreview.net/forum?id=SJ1kSyO2jwu}
}

@inproceedings{chen2023executing,
  title={Executing your commands via motion diffusion in latent space},
  author={Chen, Xin and Jiang, Biao and Liu, Wen and Huang, Zilong and Fu, Bin and Chen, Tao and Yu, Gang},
  booktitle={Proceedings of the IEEE/CVF conference on computer vision and pattern recognition},
  pages={18000--18010},
  year={2023}
}

@inproceedings{li2025simmotionedit,
  title={SimMotionEdit: Text-Based Human Motion Editing with Motion Similarity Prediction},
  author={Li, Zhengyuan and Cheng, Kai and Ghosh, Anindita and Bhattacharya, Uttaran and Gui, Liangyan and Bera, Aniket},
  booktitle={Proceedings of the Computer Vision and Pattern Recognition Conference},
  pages={27827--27837},
  year={2025}
}

@inproceedings{jiang2025dynamic,
  title={Dynamic motion blending for versatile motion editing},
  author={Jiang, Nan and Li, Hongjie and Yuan, Ziye and He, Zimo and Chen, Yixin and Liu, Tengyu and Zhu, Yixin and Huang, Siyuan},
  booktitle={Proceedings of the Computer Vision and Pattern Recognition Conference},
  pages={22735--22745},
  year={2025}
}

@inproceedings{pinyoanuntapong2024mmm,
  title={Mmm: Generative masked motion model},
  author={Pinyoanuntapong, Ekkasit and Wang, Pu and Lee, Minwoo and Chen, Chen},
  booktitle={Proceedings of the IEEE/CVF Conference on Computer Vision and Pattern Recognition},
  pages={1546--1555},
  year={2024}
}

@inproceedings{kim2023flame,
  title={Flame: Free-form language-based motion synthesis \& editing},
  author={Kim, Jihoon and Kim, Jiseob and Choi, Sungjoon},
  booktitle={Proceedings of the AAAI Conference on Artificial Intelligence},
  volume={37},
  number={7},
  pages={8255--8263},
  year={2023}
}

@inproceedings{athanasiou2023sinc,
  title={SINC: Spatial composition of 3D human motions for simultaneous action generation},
  author={Athanasiou, Nikos and Petrovich, Mathis and Black, Michael J and Varol, G{\"u}l},
  booktitle={Proceedings of the IEEE/CVF International Conference on Computer Vision},
  pages={9984--9995},
  year={2023}
}

@article{ho2016generative,
  title={Generative adversarial imitation learning},
  author={Ho, Jonathan and Ermon, Stefano},
  journal={Advances in neural information processing systems},
  volume={29},
  year={2016}
}

@article{peng2021amp,
  title={Amp: Adversarial motion priors for stylized physics-based character control},
  author={Peng, Xue Bin and Ma, Ze and Abbeel, Pieter and Levine, Sergey and Kanazawa, Angjoo},
  journal={ACM Transactions on Graphics (ToG)},
  volume={40},
  number={4},
  pages={1--20},
  year={2021},
  publisher={ACM New York, NY, USA}
}

@article{gulrajani2017improved,
  title={Improved training of wasserstein gans},
  author={Gulrajani, Ishaan and Ahmed, Faruk and Arjovsky, Martin and Dumoulin, Vincent and Courville, Aaron C},
  journal={Advances in neural information processing systems},
  volume={30},
  year={2017}
}

@article{peng2022ase,
  title={Ase: Large-scale reusable adversarial skill embeddings for physically simulated characters},
  author={Peng, Xue Bin and Guo, Yunrong and Halper, Lina and Levine, Sergey and Fidler, Sanja},
  journal={ACM Transactions On Graphics (TOG)},
  volume={41},
  number={4},
  pages={1--17},
  year={2022},
  publisher={ACM New York, NY, USA}
}

@inproceedings{tessler2023calm,
  title={Calm: Conditional adversarial latent models for directable virtual characters},
  author={Tessler, Chen and Kasten, Yoni and Guo, Yunrong and Mannor, Shie and Chechik, Gal and Peng, Xue Bin},
  booktitle={ACM SIGGRAPH 2023 Conference Proceedings},
  pages={1--9},
  year={2023}
}

@article{zhu2023neural,
  title={Neural categorical priors for physics-based character control},
  author={Zhu, Qingxu and Zhang, He and Lan, Mengting and Han, Lei},
  journal={ACM Transactions on Graphics (TOG)},
  volume={42},
  number={6},
  pages={1--16},
  year={2023},
  publisher={ACM New York, NY, USA}
}

@article{lee2022learning,
  title={Learning virtual chimeras by dynamic motion reassembly},
  author={Lee, Seyoung and Lee, Jiye and Lee, Jehee},
  journal={ACM Transactions on Graphics (TOG)},
  volume={41},
  number={6},
  pages={1--13},
  year={2022},
  publisher={ACM New York, NY, USA}
}

@article{xu2023composite,
  title={Composite motion learning with task control},
  author={Xu, Pei and Shang, Xiumin and Zordan, Victor and Karamouzas, Ioannis},
  journal={ACM Transactions on Graphics (TOG)},
  volume={42},
  number={4},
  pages={1--16},
  year={2023},
  publisher={ACM New York, NY, USA}
}

@article{van2016learning,
  title={Learning values across many orders of magnitude},
  author={Van Hasselt, Hado P and Guez, Arthur and Hessel, Matteo and Mnih, Volodymyr and Silver, David},
  journal={Advances in neural information processing systems},
  volume={29},
  year={2016}
}

@article{xu2023adaptnet,
  title={Adaptnet: Policy adaptation for physics-based character control},
  author={Xu, Pei and Xie, Kaixiang and Andrews, Sheldon and Kry, Paul G and Neff, Michael and McGuire, Morgan and Karamouzas, Ioannis and Zordan, Victor},
  journal={ACM Transactions on Graphics (TOG)},
  volume={42},
  number={6},
  pages={1--17},
  year={2023},
  publisher={ACM New York, NY, USA}
}

@article{won2022physics,
  title={Physics-based character controllers using conditional vaes},
  author={Won, Jungdam and Gopinath, Deepak and Hodgins, Jessica},
  journal={ACM Transactions on Graphics (TOG)},
  volume={41},
  number={4},
  pages={1--12},
  year={2022},
  publisher={ACM New York, NY, USA}
}

@inproceedings{pan2025tokenhsi,
  title={Tokenhsi: Unified synthesis of physical human-scene interactions through task tokenization},
  author={Pan, Liang and Yang, Zeshi and Dou, Zhiyang and Wang, Wenjia and Huang, Buzhen and Dai, Bo and Komura, Taku and Wang, Jingbo},
  booktitle={Proceedings of the Computer Vision and Pattern Recognition Conference},
  pages={5379--5391},
  year={2025}
}

@inproceedings{xu2024synchronize,
  title={Synchronize dual hands for physics-based dexterous guitar playing},
  author={Xu, Pei and Wang, Ruocheng},
  booktitle={SIGGRAPH Asia 2024 Conference Papers},
  pages={1--11},
  year={2024}
}

@inproceedings{bae2023pmp,
  title={Pmp: Learning to physically interact with environments using part-wise motion priors},
  author={Bae, Jinseok and Won, Jungdam and Lim, Donggeun and Min, Cheol-Hui and Kim, Young Min},
  booktitle={ACM SIGGRAPH 2023 Conference Proceedings},
  pages={1--10},
  year={2023}
}

@inproceedings{bae2025plt,
  title={PLT: Part-Wise Latent Tokens as Adaptable Motion Priors for Physically Simulated Characters},
  author={Bae, Jinseok and Lee, Younghwan and Lim, Donggeun and Kim, Young Min},
  booktitle={Proceedings of the Special Interest Group on Computer Graphics and Interactive Techniques Conference Conference Papers},
  pages={1--10},
  year={2025}
}

@inproceedings{khoshsiyar2024partwisempc,
  title={PartwiseMPC: Interactive Control of Contact-Guided Motions},
  author={Khoshsiyar, Niloofar and Gou, Ruiyu and Zhou, Tianhong and Andrews, Sheldon and van de Panne, M},
  booktitle={Computer Graphics Forum},
  volume={43},
  number={8},
  pages={e15174},
  year={2024},
  organization={Wiley Online Library}
}

@inproceedings{luo2023perpetual,
  title={Perpetual humanoid control for real-time simulated avatars},
  author={Luo, Zhengyi and Cao, Jinkun and Kitani, Kris and Xu, Weipeng and others},
  booktitle={Proceedings of the IEEE/CVF International Conference on Computer Vision},
  pages={10895--10904},
  year={2023}
}

@article{won2020scalable,
  title={A scalable approach to control diverse behaviors for physically simulated characters},
  author={Won, Jungdam and Gopinath, Deepak and Hodgins, Jessica},
  journal={ACM Transactions on Graphics (TOG)},
  volume={39},
  number={4},
  pages={33--1},
  year={2020},
  publisher={ACM New York, NY, USA}
}

@article{tessler2024maskedmimic,
  title={Maskedmimic: Unified physics-based character control through masked motion inpainting},
  author={Tessler, Chen and Guo, Yunrong and Nabati, Ofir and Chechik, Gal and Peng, Xue Bin},
  journal={ACM Transactions on Graphics (TOG)},
  volume={43},
  number={6},
  pages={1--21},
  year={2024},
  publisher={ACM New York, NY, USA}
}

@inproceedings{he2025hover,
  title={Hover: Versatile neural whole-body controller for humanoid robots},
  author={He, Tairan and Xiao, Wenli and Lin, Toru and Luo, Zhengyi and Xu, Zhenjia and Jiang, Zhenyu and Kautz, Jan and Liu, Changliu and Shi, Guanya and Wang, Xiaolong and others},
  booktitle={2025 IEEE International Conference on Robotics and Automation (ICRA)},
  pages={9989--9996},
  year={2025},
  organization={IEEE}
}

@article{tevet2024closd,
  title={Closd: Closing the loop between simulation and diffusion for multi-task character control},
  author={Tevet, Guy and Raab, Sigal and Cohan, Setareh and Reda, Daniele and Luo, Zhengyi and Peng, Xue Bin and Bermano, Amit H and van de Panne, Michiel},
  journal={arXiv preprint arXiv:2410.03441},
  year={2024}
}

@article{zhao2024dartcontrol,
  title={DartControl: A diffusion-based autoregressive motion model for real-time text-driven motion control},
  author={Zhao, Kaifeng and Li, Gen and Tang, Siyu},
  journal={arXiv preprint arXiv:2410.05260},
  year={2024}
}

@inproceedings{BABEL:CVPR:2021,
  title = {{BABEL}: Bodies, Action and Behavior with English Labels},
  author = {Punnakkal, Abhinanda R. and Chandrasekaran, Arjun and Athanasiou, Nikos and Quiros-Ramirez, Alejandra and Black, Michael J.},
  booktitle = {Proceedings IEEE/CVF Conf.~on Computer Vision and Pattern Recognition (CVPR)},
  pages = {722--731},
  month = jun,
  year = {2021},
  doi = {},
  month_numeric = {6}
}

@conference{AMASS:ICCV:2019,
  title = {{AMASS}: Archive of Motion Capture as Surface Shapes},
  author = {Mahmood, Naureen and Ghorbani, Nima and Troje, Nikolaus F. and Pons-Moll, Gerard and Black, Michael J.},
  booktitle = {International Conference on Computer Vision},
  pages = {5442--5451},
  month = oct,
  year = {2019},
  month_numeric = {10}
}

@article{harvey2020robust,
author    = {Félix G. Harvey and Mike Yurick and Derek Nowrouzezahrai and Christopher Pal},
title     = {Robust Motion In-Betweening},
booktitle = {ACM Transactions on Graphics (Proceedings of ACM SIGGRAPH)},
publisher = {ACM},
volume    = {39},
number    = {4},
year      = {2020}
}

@inproceedings{zhou2019continuity,
  title={On the continuity of rotation representations in neural networks},
  author={Zhou, Yi and Barnes, Connelly and Lu, Jingwan and Yang, Jimei and Li, Hao},
  booktitle={Proceedings of the IEEE/CVF conference on computer vision and pattern recognition},
  pages={5745--5753},
  year={2019}
}

@article{makoviychuk2021isaac,
  title={Isaac gym: High performance gpu-based physics simulation for robot learning},
  author={Makoviychuk, Viktor and Wawrzyniak, Lukasz and Guo, Yunrong and Lu, Michelle and Storey, Kier and Macklin, Miles and Hoeller, David and Rudin, Nikita and Allshire, Arthur and Handa, Ankur and others},
  journal={arXiv preprint arXiv:2108.10470},
  year={2021}
}

@article{schulman2017proximal,
  title={Proximal policy optimization algorithms},
  author={Schulman, John and Wolski, Filip and Dhariwal, Prafulla and Radford, Alec and Klimov, Oleg},
  journal={arXiv preprint arXiv:1707.06347},
  year={2017}
}

@inproceedings{radford2021learning,
  title={Learning transferable visual models from natural language supervision},
  author={Radford, Alec and Kim, Jong Wook and Hallacy, Chris and Ramesh, Aditya and Goh, Gabriel and Agarwal, Sandhini and Sastry, Girish and Askell, Amanda and Mishkin, Pamela and Clark, Jack and others},
  booktitle={International conference on machine learning},
  pages={8748--8763},
  year={2021},
  organization={PmLR}
}
